\documentclass[letterpaper]{article}[11pt]

\usepackage[a4paper,top=2.5cm,bottom=2.5cm,left=3cm,right=3cm,marginparwidth=1.75cm]{geometry}

\usepackage{authblk}
\usepackage{natbib}
\usepackage{times}  %
\usepackage{helvet} %
\usepackage{courier}  %
\usepackage{graphicx}
\usepackage{subcaption}
\usepackage{booktabs} %
\usepackage[hyphens]{url}
\usepackage{amsmath}
\usepackage{algorithm}
\usepackage{algorithmic}
\usepackage{amssymb}
\usepackage{bbm}
\usepackage{gensymb}
\usepackage{amsmath,amssymb}
\usepackage{balance}
\usepackage{fancyhdr}
\usepackage{color, colortbl}

\DeclareMathOperator{\E}{\mathbb{E}}
\DeclareMathOperator{\R}{\mathbb{R}}
\definecolor{LightCyan}{rgb}{0.88,1,1}
\definecolor{aqua}{rgb}{0,1,1}
\definecolor{lightgray}{rgb}{0.83, 0.83, 0.83}

\title{\textbf{Safe Option-Critic: Learning Safety in the Option-Critic Architecture}}
\author[1,2,*]{Arushi Jain}
\author[1,2,*]{Khimya Khetarpal}
\author[1,2,3]{Doina Precup}
\affil[1]{McGill University, Montreal.}
\affil[2]{Mila, Montreal.}
\affil[3]{Google DeepMind, Montreal.}
\affil[*]{\small{Equal Contribution.}}
\affil[ ]{\textit{\{arushi.jain, khimya.khetarpal\}@mail.mcgill.ca, dprecup@cs.mcgill.ca}}
\date{}

\begin{document}

\maketitle

\thispagestyle{fancy}
\renewcommand{\footrulewidth}{1pt}
\renewcommand{\headrulewidth}{0pt}
\fancyhf{}
\lfoot{To appear at The \textit{Knowledge Engineering Review} (KER), 2021. A previous draft appeared in \textit{Adaptive Learning Agents} (ALA) 2018 workshop held at ICML, AAMAS in Stockholm. }

\begin{abstract}Designing hierarchical reinforcement learning algorithms that exhibit safe behaviour is not only vital for practical applications, but also, facilitates a better understanding of an agent's decisions. We tackle this problem in the options framework (Sutton, Precup \& Singh, 1999), a particular way to specify temporally abstract actions which allow an agent to use sub-policies with start and end conditions. We consider a behaviour as \textit{safe} that avoids regions of state-space with high uncertainty in the outcomes of actions. We propose an optimization objective that learns \textit{safe} options by encouraging the agent to visit states with higher behavioural consistency. The proposed objective results in a trade-off between maximizing the standard expected return and minimizing the effect of model uncertainty in the return. We propose a policy gradient algorithm to optimize the constrained objective function. We examine the quantitative and qualitative behaviour of the proposed approach in a tabular grid-world, continuous-state puddle-world, and three games from the Arcade Learning Environment: Ms.Pacman, Amidar, and Q*Bert. Our approach achieves a reduction in the variance of return, boosts performance in environments with intrinsic variability in the reward structure, and compares favorably both with primitive actions as well as with risk-neutral options.

\end{abstract}

\section{INTRODUCTION}
\label{submission}

Safety in Artificial Intelligence (AI) has become an important focus of research as AI methods make rapid advancements~\citep{concreteproblems}.  %
The $23$ Asilomar AI principles~\citep{asilomar2017} discuss a variety of safety aspects such as risk-aversion, transparency, robustness, fairness, and also legal and ethical values that an agent must hold. In this work, we mainly focus on safety from the perspective of preventing undesirable behaviour, in particular, reducing visits to undesirable or risky states during the learning process of a reinforcement learning (RL) system.

RL algorithms optimize the expected value of cumulative rewards (return) obtained over time~\citep{Sutton:1998:IRL:551283}.
For safety-critical applications such as finance, medicine, or industrial automation, optimizing the expected return alone is often not sufficient. An agent might exhibit undesirable behaviour over certain trajectories or for a certain period. In general, safety requires measuring some notion of {\em risk} or {\em uncertainty} in addition to maximizing the mean return thus ensuring that the learning algorithm also minimizes risk.  

An additional challenge in RL is that agents need to explore during which an agent might be unaware of the states prone to noise or potentially leading to catastrophic consequences. Risk-awareness in RL has been formulated in several ways, for example by directed exploration that avoids the high entropy states \citep{law2005risk}, optimizing the worst-case performance instead of the expected performance of an agent \citep{tamar2013scaling}, measuring and bounding the probabilities of visiting the erroneous states \citep{geibel2005risk} and several other approaches. \citet{garcia2015comprehensive} present a comprehensive survey covering a broad range of techniques aimed at achieving safety in RL. In the context of Markov Decision Processes (MDPs), the source of uncertainty or variability in a sequential decision making task can arise mainly from two sources - inherent stochastic uncertainty in the reward and the transition model (aleatoric uncertainty), and imperfect knowledge regarding the model (epistemic uncertainty). The former uncertainty is usually clustered under \textit{risk-sensitive MDPs}~\citep{howard1972risk, white1994mathematical, heger1994consideration, borkar2002risk}, whereas the latter is covered under \textit{robust MDPs}~\citep{nilim2005robust, iyengar2005robust, lim2013reinforcement}. In this work, we focus on risk-sensitive MDPs and address the former source of variability via mean-variance optimization.

The natural question that follows is how can we measure variance in RL? Several works focus on estimating the variance in return using temporal difference (TD) style learning methods~\citep{tamar2012policy, tamar2016learning, Gehring:2013, sherstan2018directly, jain2021variance}. Our work is based on the specific approach proposed by \cite{Gehring:2013}, which suggests using the absolute value of the TD error to define \textit{controllability} of a state. Intuitively, for an agent to exhibit safe behaviour, it will prefer taking the actions whose effects are more predictable on a state. Such states are referred as \emph{controllable}. We re-define controllability slightly to provide an estimator which calculates the uncertainty in the value function, and then show how to ``encourage" an agent to focus on controllable states. %

All approaches discussed so far define safety in the primitive action space (one step action). Inspired by how humans think and plan, temporally abstract actions also provide a path to learn and plan efficiently. Temporal abstractions have been a crucial part of AI research since the $1970$s \citep{fikes1972learning, fikes1981learning, iba1989heuristic, korf1983learning, parr1998reinforcement, SUTTON1999181, precup2000temporal, dietterich2000hierarchical, mcgovern2001automatic, menache2002q, barto2003recent, konidaris2007building, bacon2017option, barreto2019option}. Prior research has shown that the temporal abstractions can improve exploration, reduce the complexity of deliberation and enhance robustness. More importantly, from the safety perspective, temporal abstractions lead to composition of  behaviour and predictions. To simplify, if every subpart of the abstraction hierarchy can be derived respecting the safety conditions, the entire hierarchy would automatically be safe in behaviour. In this paper, we take the first step in this direction by showing how to construct temporally safe extended actions by reducing the variability or uncertainty as discussed above.

We approach this problem in the \textit{options} framework \citep{SUTTON1999181, precup2000temporal}, which provides an intuitive way to plan, reason, and act at multiple timescales. We are especially interested in learning options end-to-end from the data obtained by the agent's interaction with the environment. The options framework has received a lot of attention over time, for example, \cite{stolle2002learning, daniel2016probabilistic, konidaris2007building, konidaris2011autonomous, kulkarni2016hierarchical, vezhnevets2016strategic,mankowitz2016adaptive, jain2018eligibility, riemer2018learning, barreto2019option}. 

\cite{bacon2017option} introduced the option-critic framework by adapting to the actor-critic model. The option-critic model facilitates an end-to-end learning of options without the need to specify sub-goals. We modify the option-critic objective to include the negation of controllability \citep{Gehring:2013} (estimates the variability in the performance) as a regularizer in the optimization objective. Consequently, the \textbf{key contributions} of this work can be summarised as follows:
\begin{enumerate}
    \item We propose a new objective function which uses the effect of the uncertainty in the return as a regularizer for the classical optimization problem of maximizing the expected return. We then use this objective to derive a policy gradient algorithm for automatically learning safe options.
    \item We empirically demonstrate the effectiveness of our approach in the tabular setup.
    \item We also show that our method is scalable to both linear and non-linear function approximation settings using the puddle-world and ALE games: Ms.Pacman, Amidar, and Q*Bert.
\end{enumerate}

The paper is structured as follows. After presenting definitions and notations in Sec.~\ref{sec:preliminaries}, we introduce the safe option-critic algorithm in Sec.~\ref{sec:socmethod}. Next, we present tabular, continuous and ALE experiments in Sec.~\ref{sec:experiments}. Finally, we conclude with discussion and future work in Sec.~\ref{sec:conclusionandfuturework}.

\section{PRELIMINARIES}
\label{sec:preliminaries}
In RL, an agent interacts with the environment at discrete time steps $t \in \{1,2,...\}$. At each step, the agent observes a state $s \in \mathcal{S}$ and chooses an action $a \in \mathcal{A}$ according to a policy which defines a probability distribution of actions over the state space $\pi: \mathcal{S} \times \mathcal{A} \to [0,1]$. Afterwards, the agent transitions from $S_t$ to $S_{t+1}$ state according to the transition probability distribution $P:\mathcal{S} \times \mathcal{A} \times \mathcal{S} \to [0,1]$. After transitioning to a next state, an agent receives a reward $R_{t+1} \in \R$, where the expected reward function is $r(s,a)= \sum_{r\in \mathbbm{R}}r \sum_{s'}P(s',r|s,a)$, and $r: \mathcal{S} \times \mathcal{A} \to \mathbb{R}$. The \textit{Markov Decision Process} (MDP) is defined as a tuple of $(\mathcal{S}, \mathcal{A}, r, P, \gamma)$,  where $\gamma \in [0,1]$ is the discount factor, which de-values rewards received farther into the future. A Markov Decision Process with a defined optimality criteria is known as a \textit{Markov Decision Problem}. In this work, we focus on MDP with an optimality criteria as the \textit{expected total discounted reward}. According to this optimality criteria, we define the state-action value function as $Q(s,a) = \E_\pi[ \sum_{t=0}^{\infty} \gamma^t R_{t+1}|S_t = s, A_t = a]$. An estimate of $Q$ can be learned in an incremental fashion using temporal-difference (TD) learning methods \citep{sutton1988learning}. For example, in $TD(0)$, the agent computes a temporal difference error, which captures the difference in the agent's value estimate at two consecutive time steps:
\[\delta_t = R_{t+1} + \gamma Q(S_{t+1}, A_{t+1}) - Q(S_t, A_t).\]
The state-action value function is then updated as:\[Q(S_{t}, A_t) \gets Q(S_t,A_t) + \alpha \delta_t,\] where $\alpha\in [0,1]$ is the step size.

The policy gradient theorem \citep{sutton2000policy} provides a way to update a parameterized policy in the direction of the gradient of the expected return. Here, return is the total discounted rewards received in a trajectory. The performance of a policy $\pi$ is measured by the expected return as
\[\rho(\pi, s_0) = \E[\sum_{t=0}^{\infty}\gamma^{t} R_{t+1}| \pi, s_0],\]
where, the initial state is denoted by $s_0$. The gradient of the performance with respect to the policy parameter $\theta$ is given by:
\begin{equation}
	\frac{\partial \rho(\pi, s_0)}{\partial \theta} = \sum_{s} d^\pi (s|s_0) \sum_{a} \frac{\partial \pi(s,a)}{\partial \theta} Q_\pi(s,a),
\end{equation}
where, $d^\pi(s|s_0) = \sum_{t=0}^{\infty} \gamma^t \it{P(S_t = s | s_0, \pi)}$ is the discounted weighting of the states.

\subsection{OPTIONS}
The options framework \citep{SUTTON1999181, precup2000temporal} facilitates the incorporation of temporally abstract knowledge into RL. An option $w\in \mathcal{W}$, is defined as a tuple of $(I_w, \pi_w, \beta_w)$. $I_w$ is an initiation set containing initial states from which an option $w$ can start. $\pi_w$ denotes an option policy that defines a distribution over actions given a state during the execution of option $w$. The termination condition of $w$ option, $\beta_w$, defines a probability of termination in each state. 

If the option policies are Markovian, the intra-option Bellman equation \citep{SUTTON1999181} provides an off-policy method for updating the value of a state-option pair as
\begin{align*}
		Q(S_t,W_t) \gets Q(S_t, W_t) + \alpha \Big(&R_{t+1} +\gamma(1-\beta_{W_t}(S_{t+1}))Q(S_{t+1}, W_t) \\
		&+\gamma \beta_{W_t}(S_{t+1})\max_{w' \in \mathcal{W}} Q(S_{t+1}, w') - Q(S_t, W_t)\Big),
\end{align*}
where, the same option $w$ continues with $1-\beta_w$ probability, or, terminates with $\beta_w$ probability and selects the next best option.

\subsection{OPTION-CRITIC ARCHITECTURE}
The above described intra-option value learning \citep{SUTTON1999181} also lays the foundation for how the options are learnt in the \textit{Option-Critic} architecture \citep{bacon2017option}. The option-critic method is a policy gradient based method for learning intra-options policies and option's termination conditions. \cite{bacon2017option} considers the call-and-return option execution model, where an option $w$ is chosen according to a policy over options $\pi_{W}$, and an intra-option policy $\pi_w$ is followed until the termination condition $\beta_w$ is met. Once the current option terminates, another option is selected using $\pi_{W}$ and the process continues. Let $\pi_{w, \theta}$ denotes intra-option policy parameterized by $\theta$, and $\beta_{w, \nu}$ represents the option termination parameterized by $\nu$. The value of executing an action $a$ at a particular state-option pair is given by $Q_U: \mathcal{S} \times \mathcal{W} \times \mathcal{A} \to \mathbb{R}$ which is defined as 
\begin{equation}\label{eqQ_U}
Q_U(s, w, a) = r(s, a) + \gamma \, \sum_{s'} P(s'| s, a) U(s',w),
\end{equation}
where, $U(s',w)$ represents the value of arriving through an option  $w$ at a state $s'$. Here, either the option $w$ could terminate with $\beta_{w, \nu}$ probability and select the other options, or could continue with the same option with $(1- \beta_{w, \nu})$ probability,
\begin{equation}\label{eq_U}
U(s',w) = (1 - \beta_{w, v} (s')) Q_{W}(s', w)\, +\, \beta_{w, v} (s') V_W(s').
\end{equation}
Here, $Q_W$ represents the value function over options, which is given by-  
\[Q_W(s,w) = \sum_a \pi_{w, \theta}(a|s) Q_{U}(s,w, a).\] $V_W$ represents the value function of executing policy over options $\pi_W$, given by- \[V_W(s) = \sum_w\pi_W(w|s) Q_{W}(s, w).\] \cite{bacon2017option} derived the gradient of the discounted return from an initial condition $(s_0, w_0) \in (\mathcal{S \times W})$ with respect to $\theta$ (intra-option policy parameter) as:
\begin{equation}
\begin{split}
	\frac{\partial \rho(\pi, s_0, w_0)}{\partial \theta} =&\sum_{s,w} \mu(s, w| s_0, w_0) \sum_{a} \frac{\partial \pi_{w, \theta}(a | s)}{\partial \theta} Q_U(s, w, a),
\end{split}
\end{equation}
where, $\mu(s, w| s_0, w_0) = \sum_{t=0}^{\infty} \gamma^t P(S_t=s,W_t=w|s_0,w_0)$ is the discounted weighting of a state-option pair. The gradient of the expected discounted return with respect to the option termination parameter $\nu$ and the initial condition $(s_1, w_0)$ is given by:
\begin{equation}\label{eq_J_with_nu}
\frac{\partial \rho(\pi, s_1, w_0)}{\partial v}=-\sum_{s',w} \mu(s', w|s_1, w_0)\frac{\partial \beta_{w, \nu}(s')}{\partial \nu} A_W(s', w),
\end{equation}
where, $A_W$ is the advantage function $A_W(s,w) = Q_W(s,w) - V_W(s)$. In \eqref{eq_J_with_nu}, the initial condition changed from $(s_0,w_0)$ to $(s_1, w_0)$ because given an initial state-option $(s_0, w_0)$, the first action is selected according to $\pi_{w_0}$ policy, and then the agent transitions to the next state $s_1$.  The updates to the termination parameter $\nu$ are made thereafter based on the advantage of switching to the next option in state $s'$.

\section{SAFE OPTION-CRITIC MODEL}
\label{sec:socmethod}
\label{derivation}
In this work, we introduce a notion of learning safe hierarchical policies using a penalty based on the variance in TD error. Previous works in the literature had focused on the measures derived from the TD-based methods for quantifying uncertainty in the value of a state or a state-action pair.

As mentioned earlier, here, we focus on addressing uncertainty arising due to inherent stochasticity present in the model (aleatoric uncertainty). The uncertainty present in the model is reflected in the form of variability in the value function. One such feasible approach to capture the uncertainty/variability in the value function would be to estimate the variance in TD error. After the policy has converged, the expected TD error would approach zero value. The variance in the TD error results from the stochasticity in the environment's reward function and the transition dynamics. We define ``safety'' as constraining the uncertainty in the value function. Therefore, we modify the objective function of maximizing the mean return by adding a regularizer on uncertainty in the value function.

Inspired by \cite{Gehring:2013}, we define {\em controllability} as a negation of the variance in TD error of a state-option action pair. We use the aforementioned definition of controllability to introduce the concept of safety in the option-critic architecture, which aids in measuring the uncertainty in the value of a state-option pair. The key idea is - the higher the variance in TD error of a state-option pair, the higher would be the uncertainty in the value function. 

In safety-critical applications, the agent must learn to avoid such state-option pairs as they induce variability in return eventually. We optimize for obtaining the maximum expected discounted return and controllability value of the \textit{initial state-option pair}. Depending on the nature of the application, one can limit or encourage the agent's visit to a state-option pair based on the degree of controllability. Introducing controllability via the TD error facilitates the linear scalability of the proposed approach with the number of state-option pairs.

Analogous to the notations used in \cite{bacon2017option}, we introduce a parameter vector described by $\Theta = [\theta, \nu]$, where, $\theta$ is an intra-option policy parameter, and $\nu$ is an option termination parameter. We assume that all options can be initialized from any state such that $I_w= \mathcal{S}, \, \forall w\in \mathcal{W}$. Uncertainty in the value of a state-option pair is measured using controllability $C$, which is given by the negation of the variance in the TD error $(\delta)$. The TD error is -
\begin{equation}\label{eq_delta}
\delta(S_t, W_t, A_t) = R_{t+1} \, + \gamma \sum_{s'} P(s' | S_t, A_t) U_\Theta(s',W_t) - Q_{U,\Theta}(S_t, W_t, A_t),
\end{equation}
where, $Q_{U,\Theta}(s,w,a)$ and $U_\Theta(s, w)$ are defined in \eqref{eqQ_U} and \eqref{eq_U} respectively. For the ease of notation, we will use $\delta_t$ to denote $\delta(S_t, W_t, A_t)$. Whenever it is not clear from the context, we will use the full notation for the TD error. The expected value of the TD error would converge to zero. Therefore, the variance in TD error can be simplified to denote controllability as:
\begin{equation}\label{eq_C}
\begin{split}
C_{\Theta}(s, w) &= - Var_{\pi_{w,\theta}}[\delta_t | S_t=s, W_t= w]\\
&= - \Big(\E_{\pi_{w,\theta}}\big[\delta_t^2 | S_t=s, W_t= w\big] - \E_{\pi_{w,\theta}}\big[\delta_t | S_t=s, W_t= w\big]^2\Big)\\
&= -\E_{\pi_{w,\theta}}\big[\delta_t^2 | S_t=s, W_t= w\big].    
\end{split}
\end{equation}
 The aim here is to maximize both the expected discounted return and the controllability value from an initial state-option pair. We want to maximize the objective function $J$,
\begin{equation}\label{eq_J}
\begin{split}
&\max_\Theta J(\Theta | \kappa),
\\
&\text{where, }\, J(\Theta | \kappa) = \E_{(s_0, w_0) \sim \kappa}\Big[Q_\Theta(s_0, w_0) + \psi\, C_\Theta(s_0, w_0)\Big],
\end{split}
\end{equation}
$\psi \in \mathbb{R}$ is a regularizer on the controllability value, and $\kappa$ denotes the initial state-option pair distribution. The value of a state-option pair is defined as $Q_\Theta(s, w) = \sum_{a} \pi_{w, \theta} (a | s)\,  Q_{U, \Theta}(s, w, a)$. Another interpretation is to view the above objective as a constrained optimization problem, where, the constraints are imposed on the controllability function. One could bound the controllability value, but, it would require specific assumptions about the environment or one needs to rely on an input value from the user. In this paper, we opted to keep the method general without the need to rely on certain assumptions regarding the environment. We will now derive the gradient of the performance evaluator $J$ with respect to the intra-option policy parameter $\theta$, assuming they are differentiable.

First, we calculate the gradient of the $\delta_t$ with $\theta$ using \eqref{eq_delta}:
\begin{equation}\label{td_grad}
\begin{split}
    \frac{\partial \delta_t}{\partial \theta} &= \gamma \sum_{s'} P(s'| S_t, A_t)\frac{\partial U_\Theta(s', W_t)}{\partial \theta} - \frac{\partial Q_{U, \Theta}(S_t, W_t, A_t)}{\partial \theta}\\
    &= \gamma \sum_{s'} P(s'| S_t, A_t)\frac{\partial U_\Theta(s', W_t)}{\partial \theta} - \Big(\gamma \sum_{s'} P(s'| S_t, A_t)\frac{\partial U_\Theta(s', W_t)}{\partial \theta}\Big)\\
    &=0
\end{split}
\end{equation}
The above equation follows through \eqref{eqQ_U}. Next, we take the gradient of $C$ with respect to $\theta$. Following from \eqref{eq_C},
\begin{equation}\label{grad_C}
\begin{split}
\frac{\partial C_\Theta(s_0, w_0)}{\partial \theta} &= -\frac{\partial \{\sum_a \pi_{w_0, \theta}(a | s_0) \delta(s_0, w_0, a)^2\}}{\partial \theta}
\\ &=-\sum_{a}\frac{\partial \pi_{w_0, \theta}(a | s_0)}{\partial \theta}\, \delta(s_0, w_0, a)^2 -\,  2\sum_{a}\pi_{w_0, \theta}(a | s_0) \delta(s_0, w_0, a) \frac{\partial \delta(s_0, w_0, a)}{\partial \theta}\\
&=-\sum_{a}\frac{\partial \pi_{w_0, \theta}(a | s_0)}{\partial \theta}\, \delta(s_0, w_0, a)^2 \qquad\text{[Using \eqref{td_grad}]}
\end{split}
\end{equation}
The last term in the above equation vanishes because the gradient of TD error with respect to the policy parameter $\theta$ is zero (using \eqref{td_grad}).

The 1-step state-option transition is denoted by
\[P_{\gamma}^{(1)}(s', w' | s, w) = \gamma\sum_{a}\pi_{w,\theta}(a|s)P(s'|s, a)\big[\, (1-\beta_{w,\nu}(s'))\mathbbm{1}_{w= w'} + \beta_{w,\nu}(s')\pi_W(w' | s')\big].\]
Using the above 1-step transition, similarly, the k-step state-option transition is expressed as \[P_{\gamma}^{(k)}(s', w' | s_0, w_0) = P_{\gamma}^{(1)}(s_1, w_1 | s_0, w_0) \times P_{\gamma}^{(k-1)}(s', w' | s_1, w_1).\]
Using the \textit{Intra-Option Policy Gradient Theorem} \citep{bacon2017option}, the gradient of $Q_\Theta(s,w)$ with $\theta$ is: 
\begin{align}\label{intra_option_gradient}
\frac{\partial Q_\Theta(s_0,w_0)}{\partial \theta} = \sum_{k=0}^{\infty} \sum_{s',w'}P_{\gamma}^{(k)}(s',w' | s_0,w_0) \sum_{a} \frac{\partial \pi_{w', \theta}(a | s')}{\partial \theta} Q_{U, \Theta}(s',w', a).
\end{align}
Following the objective function \eqref{eq_J}, combining the gradient of $Q_\Theta(s_0,w_0)$  \eqref{intra_option_gradient} and $C_\Theta(s_0,w_0)$ \eqref{grad_C}, the gradient of $J$ with $\theta$ is:
\begin{equation}\label{j_grad_theta}
\begin{split}
   \frac{\partial J(\Theta| \kappa)}{\partial \theta} =& \sum_{k=0}^{\infty}\sum_{s, w} \Big\{P_\gamma^{(k)} (s, w| s_0, w_0) \sum_{a} \frac{\partial \pi_{w, \theta}(a |s)}{\partial \theta} Q_{U, \Theta}(s, w, a) \Big\} \\
& - \psi\sum_{a} \frac{\partial \pi_{w_0, \theta}(a |s_0)}{\partial \theta} \delta(s_0,w_0,a)^2
\end{split}
\end{equation}

In \eqref{j_grad_theta}, $(s_0,w_0)$ corresponds to the initial state-option pair. In the above equation, one can simply replace $Q_{U, \Theta}(s, w, a)$ with the advantage term $A_{\Theta}(s,w,a) = Q_{U, \Theta}(s, w, a) - Q_{\Theta}(s,w)$ because, 
\begin{equation}
\sum_{a} \frac{\partial \pi_{w, \theta}(a |s)}{\partial \theta} Q_{\Theta}(s,w) = Q_{\Theta}(s,w) \frac{\partial \sum_{a} \pi_{w, \theta}(a |s)}{\partial \theta} = Q_{\Theta}(s,w) \frac{\partial 1}{\partial \theta}  = 0.
\end{equation}
The gradient update of $J$ here describes that each option aims to maximize its own reward, along with, maintaining a high controllability value pertaining to that option policy only. 

Now, we will compute the gradient of $J(\Theta | \kappa)$ with respect to the option termination function parameter $\nu$. First, we would calculate the gradient of TD error with $\nu$.
\begin{equation}\label{Td_nu_grad}
    \begin{split}
        \frac{\partial \delta_t}{\partial v} =& \gamma \sum_{s'} P(s' | S_t, A_t) \frac{\partial U_\Theta(s', W_t)}{\partial \nu} - \frac{\partial Q_{U,\Theta}(S_t, W_t, A_t)}{\partial \nu} \\
        &=\gamma \sum_{s'} P(s' | S_t, A_t) \frac{\partial U_\Theta(s', W_t)}{\partial \nu} - \Big(\gamma \sum_{s'} P(s' | S_t, A_t) \frac{\partial U_\Theta(s', W_t)}{\partial \nu} \Big)\\
        &=0
    \end{split}
\end{equation}
The last equation follows through \eqref{eqQ_U}. The gradient of controllability $C$ with $\nu$ using \eqref{eq_C} and \eqref{Td_nu_grad} can be expressed as: 
\begin{equation}\label{grad_C_v}
\begin{split}
\frac{\partial C_\Theta(s_0, w_0)}{\partial \nu} &= - \sum_a \pi_{w_0, \theta}(a | s_0)\frac{\partial \delta(s_0, w_0, a)^2}{\partial \nu}
\\ &=-2 \sum_{a} \pi_{w_0, \theta}(a | s_0)\, \delta(s_0, w_0, a) \frac{\partial \delta(s_0, w_0, a)}{\partial \nu}
\\ &= 0 \qquad \text{[Using \eqref{Td_nu_grad}]}
\end{split}
\end{equation}
Therefore, the gradient of $J$ with $\nu$ remains same as the gradient of the state-option value. On using the \textit{Termination Gradient Theorem} \citep{bacon2017option}, the gradient of $J$ with $\nu$ results as follows:
\begin{equation}\label{eq_J_v_f}
\begin{split}
\frac{\partial J(\Theta| \kappa)}{\partial \nu} &= \frac{\partial Q_\Theta(s_1, w_0)}{\partial \nu} \\
&= - \sum_{k=0}^{\infty}\sum_{s, w} P_\gamma^{(k)} (s, w| s_1, w_0) \frac{\partial \beta_{w,\nu}(s)}{\partial \theta} A_{W}(s, w).
\end{split}
\end{equation}
Our interpretation of the above derivation follows the notion of safety; each option is responsible for making its intra-option policy safe by incorporating the controllability function. Due to the assumption that each option takes care of its own safety through the intra-option policy (the objective function implicitly does not bring changes at the policy over options level), one is only concerned about choosing an option that maximizes the expected discounted return from the next state-option pair while terminating an option. As a result, introduction of controllability does not impact the termination function update. Algorithm~\ref{alg:safe-oc} shows a prototype implementation details of controllability in the option-critic architecture in the look-up table setting.

We now provide some intuition behind using the variance of TD error to learn the controllable states. The variance in the TD error of the initial state-option-action $\delta(s_0,w_0, a_0)$ uses the difference in the current state-option value $Q_{U,\Theta}(S_t, W_t, A_t)$, and the next expected state-option value $\E[U_{\Theta}(s', W_t)| S_t =s]$. With the increase in the time steps, the estimate of the Q value function changes (on seeing the new reward samples along a trajectory, the Q function estimate changes). Therefore, the square of the TD error would capture the variations in the return (sum of rewards) from the initial state-option pair on wards, thus yielding the uncertainty in the value function estimates.

{
	\begin{algorithm}[t]
		\caption{Safe Option-Critic with look-up table intra-option Q learning}
		\label{alg:safe-oc}
		\begin{algorithmic}
			\STATE Here $\alpha_c, \alpha_\theta, \alpha_\nu$ stands for the step size of critic, intra-option policy and termination respectively. $\psi$ is controllability regularization parameter. 
			
			\STATE $s \gets s_0$
			\STATE Take $w\sim \pi_W(w| s_0))$
			\STATE Let initial $w$ be $w_0$
			\REPEAT
			\STATE Take $a \sim \pi_{w,\theta}(.|s)$
			\STATE Let initial $a$ taken at $(s_0, w_0)$ be $a_0$
			\STATE Maintain $(s_0,w_0, a_0)$ at the beginning of the episode 
			\STATE Observe $\{r, s'\}$
			\IF{$s'$ is a non-terminal state}
			\STATE $\delta \gets \, r + \gamma\Big[(1-\beta_{w, \nu}(s'))Q_{\Theta}(s',w) +\, \beta_{w,\nu}(s')\max_{w' \sim \mathcal{W}}Q_{\Theta}(s',w')\Big]-\, Q_{U,\Theta}(s, w, a)$ 
			\ELSE
			\STATE $\delta \gets r - Q_{U,\Theta}(s, w, a)$
			\ENDIF
			\STATE Maintain $\hat\delta \gets \delta^2$ at the beginning of episode with $(s_0,w_0,a_0)$
			\IF {$(s_0,w_0) == (s,w)$}
			\STATE Update $(s_0,w_0,a_0) \gets (s_0,w_0,a)$
			\STATE Update $\hat \delta \gets \delta^2$ with new $(s_0,w_0,a_0)$
			\ENDIF 
			
			\STATE $Q_{U,\Theta}(s, w, a) \gets Q_{U,\Theta}(s, w, a) + \alpha \delta$
			\STATE $\theta \gets \theta + \alpha_\theta \Big\{ \frac{\partial \log(\pi_{w, \theta}(a|s))}{\partial \theta} \big(Q_{U, \Theta}(s, w, a) - Q_\Theta(s, w)\big)- \textcolor{blue}{\psi \frac{\partial \log(\pi_{w_0, \theta}(a_0|s_0))}{\partial \theta} \hat\delta} \Big\}$
			\STATE $\nu \gets \nu - \alpha_\nu \frac{\partial \beta_{w, \nu}(s')}{\partial \nu} (Q_\Theta(s', w) - V_{W}(s'))$
			\IF{$\beta_{w, \nu}(s')$ terminates}
			\STATE Choose new $w\sim \pi_W(w| s')$
			\ENDIF
			\STATE $s \gets s'$
			\UNTIL{$s'$ is a terminal state}
		\end{algorithmic}
	\end{algorithm}
}

\section{EXPERIMENTS}
\label{sec:experiments}
\subsection{Grid World}

\begin{figure}[ht]
	\begin{center}
		\centerline{\includegraphics[width=0.28\columnwidth]{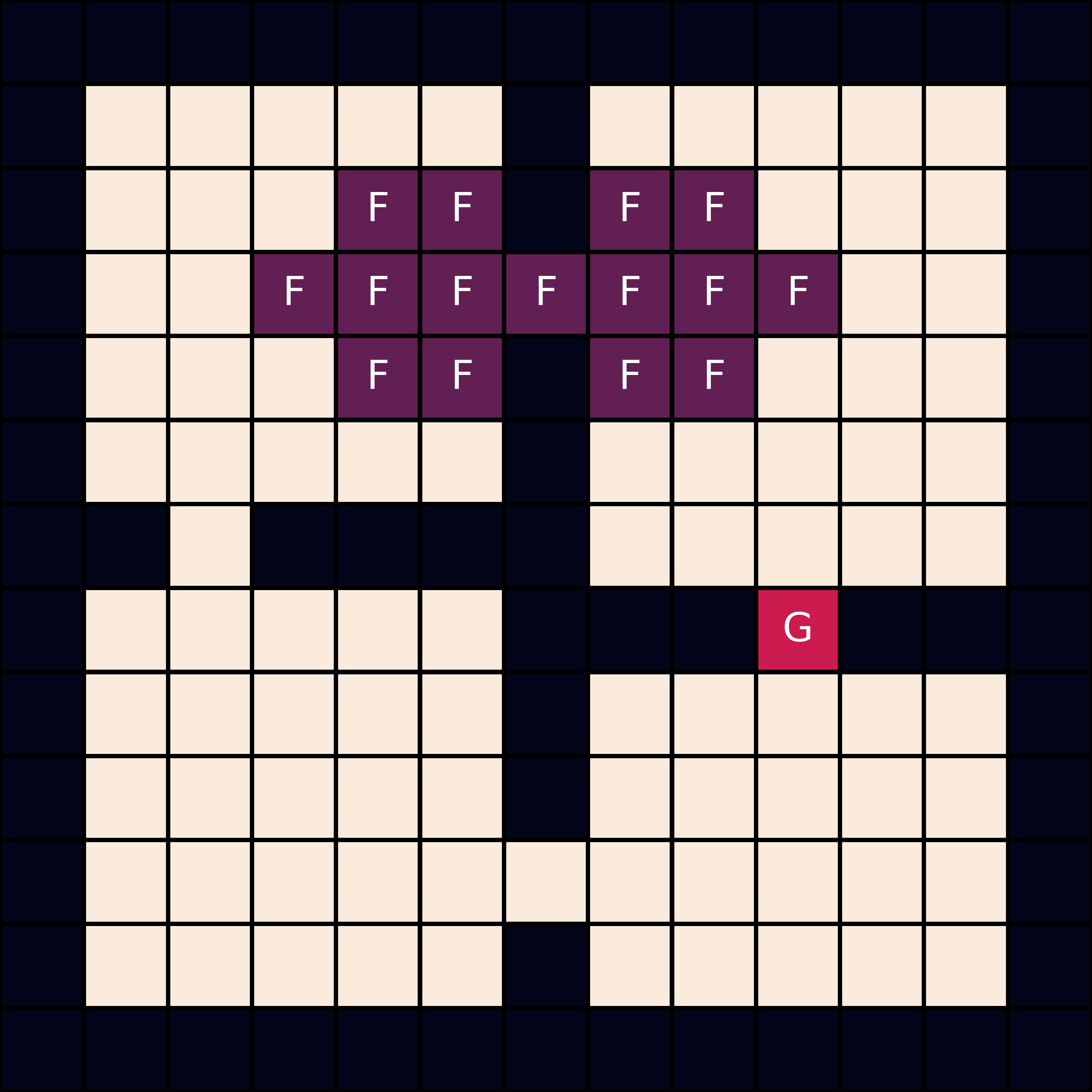}}
		\caption{\textbf{Four Room (FR) Environment}: $F$ and $G$ depicts the unsafe frozen and goal states respectively. The lightest color represents the normal states, whereas, the darkest color represents the wall.}
		\label{fig:fourroom}
	\end{center}
\end{figure}

First, we consider a simple navigation task in a two-dimensional grid environment using a variant of the four-rooms domain as described in \cite{SUTTON1999181}. As shown in Fig.~\ref{fig:fourroom}, similar to \cite{Gehring:2013}, we define some \textit{slippery} frozen states in the environment which are unsafe to visit. We accomplish this by introducing variability in the rewards of these frozen states. The states labeled as \textit{F} and \textit{G} indicate the frozen and the goal states respectively.

An agent can be initialized with any random start state in the environment apart from the goal state. The action space consists of four stochastic actions, namely, \textit{up}, \textit{down}, \textit{left}, and \textit{right}. Upon choosing an action, with $0.2$ probability, the agent can transition into any of the four directions (irrespective of the selected action). Whereas,  with $0.8$ probability, an agent takes the intended action deterministically. The task is to navigate through the rooms to a fixed goal state as depicted in Fig.~\ref{fig:fourroom}. The dark states in Fig.~\ref{fig:fourroom} depict the walls. The agent remains in the same state with a reward of $0$ if the agent hits the wall. A reward of $0$ and $50$ is given to the agent on transitioning into the normal and the goal state respectively. Rewards for the unsafe states are drawn from $\mathcal{N}(\mu=0, \sigma = 15)$ distribution when the agent transitions to a slippery state. The expected value of the reward for the normal and the slippery states is the same.

Here we are using a look-up table representation to learn the policy. In the safe option-critic framework, we represent both policy over options and intra-option policies with the Boltzmann distribution. Sigmoid activation is used for learning the termination function. We ran the experiments with different controllability factor $\psi$ for learning $4$ options. One could choose a different number of options as well. %
We optimize for the hyperparameters: temperature and step size for both the baseline Option-Critic (OC) with $\psi=0$ and Safe-OC. The discount factor $\gamma$ is set to $0.99$. Hyperparameters for the experiment are mentioned in Table \ref{tab:param-OC}. We ran the experiments for a total of $500$ episodes averaged over $50$ trials, where, the training in each trial starts from scratch. In each episode, the agent is allowed to take a maximum of only $500$ steps, wherein if the agent fails to reach the goal state within those steps, then the episode terminates.

\begin{figure}[t]
	\begin{center}
		\begin{subfigure}[b]{0.47\textwidth}
			\captionsetup{justification=centering}            \includegraphics[width=\textwidth]{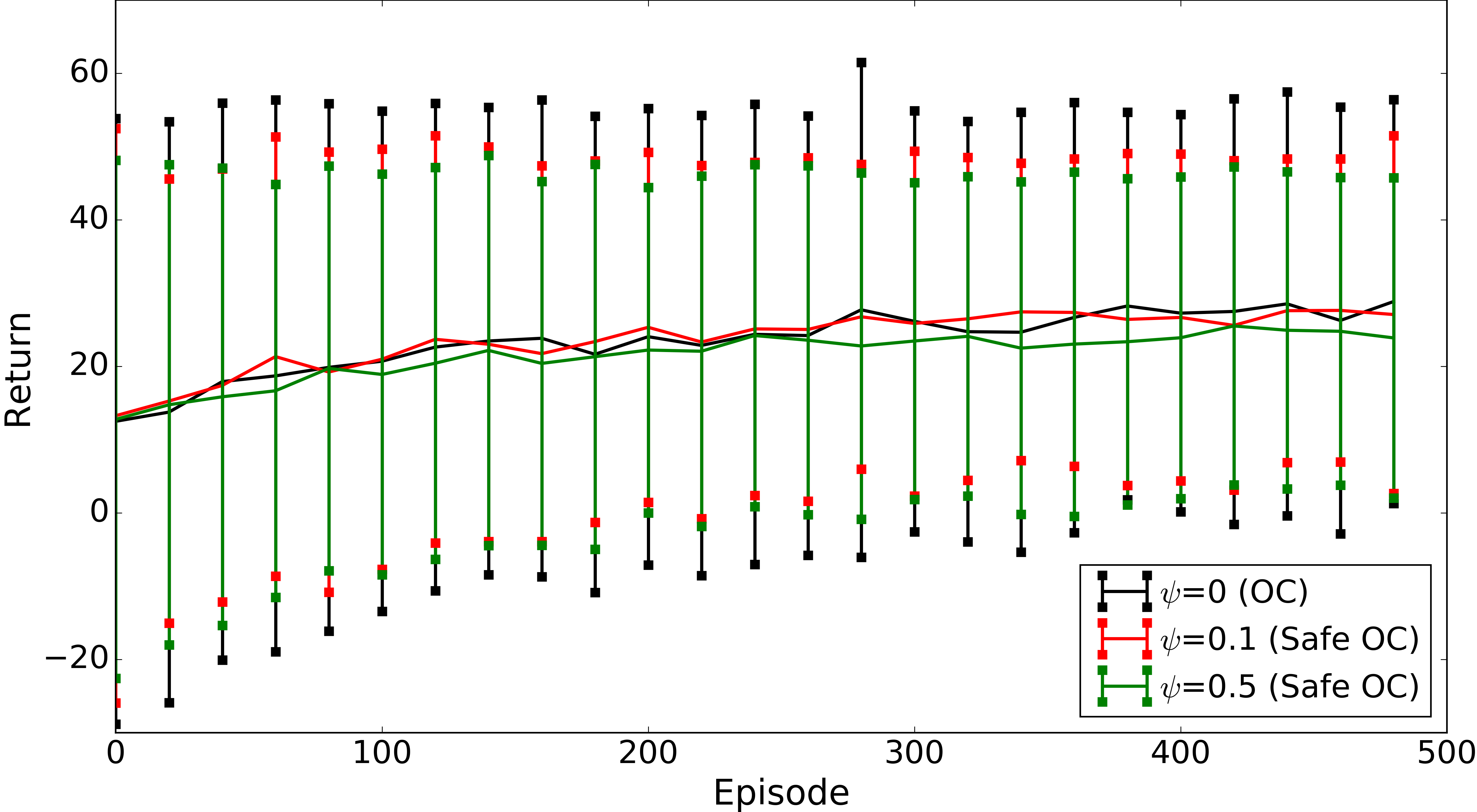}
			\caption[]%
			{{Learning Curve}}
			\label{fig:FR_return}
		\end{subfigure}
		\quad
		\begin{subfigure}[b]{0.47\textwidth}  
			\captionsetup{justification=centering}            \includegraphics[width=\textwidth]{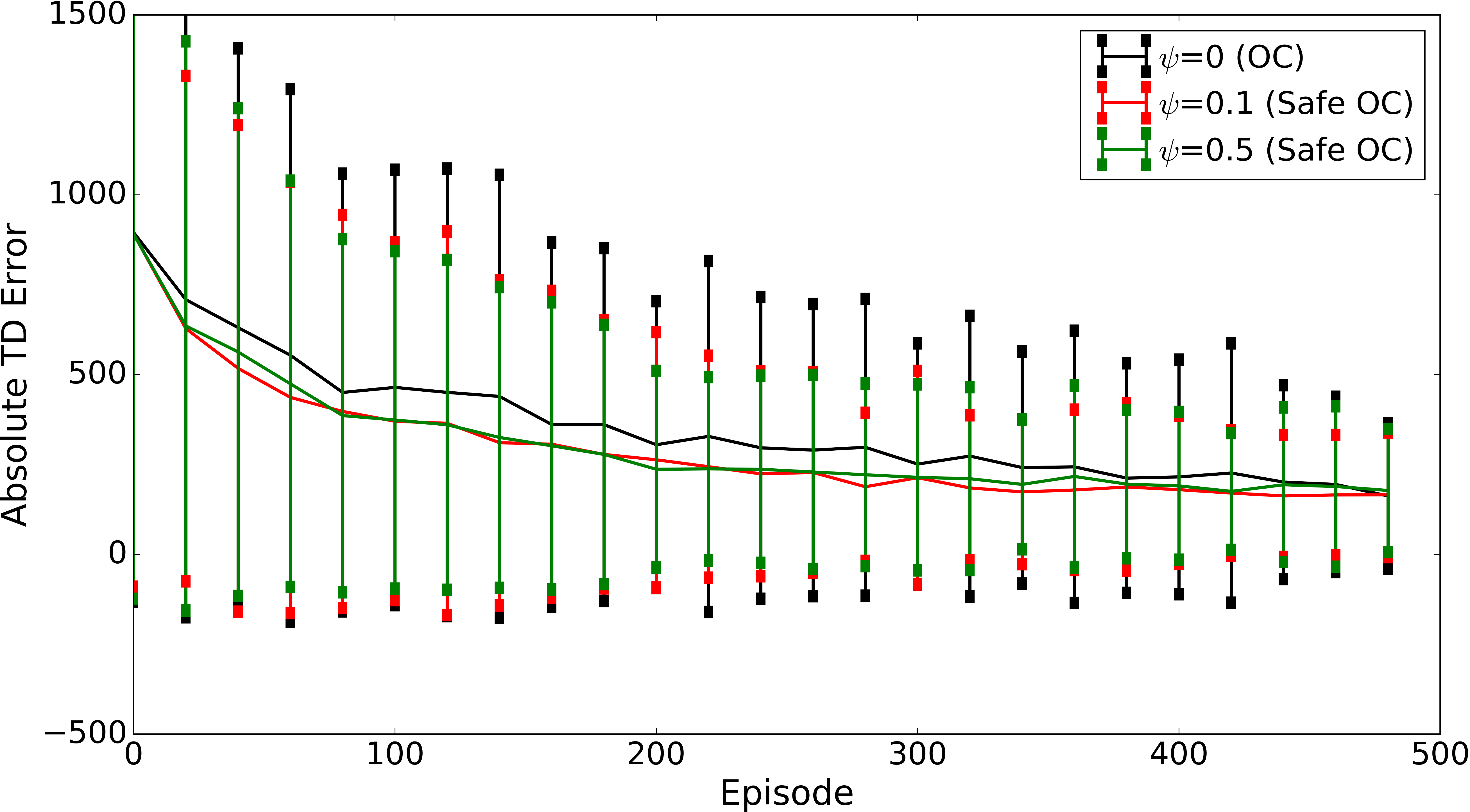}
			\caption[]%
			{{Absolute TD error}}
			\label{fig:FR_TD_error}
		\end{subfigure}
		\caption{\textbf{Performance in the FR domain.} The graphs depict the performance averaged over $50$ independent trials, where, the vertical bands depict the standard deviation. The plots show a) the return and b) the sum of absolute TD error. Safe policy $\psi=0.1$ (red) has a smaller standard deviation as compared to the baseline (black) signifying that safety helps the agent to avoid the variance inducing regions.}
		\label{fig:FourRoom4Opt}
	\end{center}        
\end{figure}

\begin{figure}[t]
	\begin{center}
		\centerline{\includegraphics[width=0.4\columnwidth]{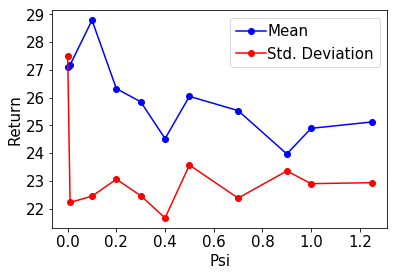}}
		\caption{\textbf{Performance with varying variance regularizer in FR domain.} The plot shows the mean and the standard deviation of the score in the Four Rooms Domain with varying controllability regularizer value $\psi$, while keeping all other parameters fixed. The best hyper-parameters are chosen such that they minimize the standard deviation, but also, maximize the mean performance, which in this case is achieved by $\psi=0.1$.}
		\label{fig:variance_regularizer_ fourroom}
	\end{center}
\end{figure}

To evaluate these experiments, we consider the following metrics: compared the expectation and the variance in the performance over multiple trials (to comment on the stability of the introduced algorithm); sum of the absolute TD error over episodes; density of the state visits. We also show sampled trajectories from both the baseline and the proposed method to qualitatively express the differences between the two after learning has been completed.

\begin{figure}[t]
	\begin{center}
		\begin{subfigure}[b]{0.25\textwidth}
			\captionsetup{justification=centering}            \includegraphics[width=\textwidth]{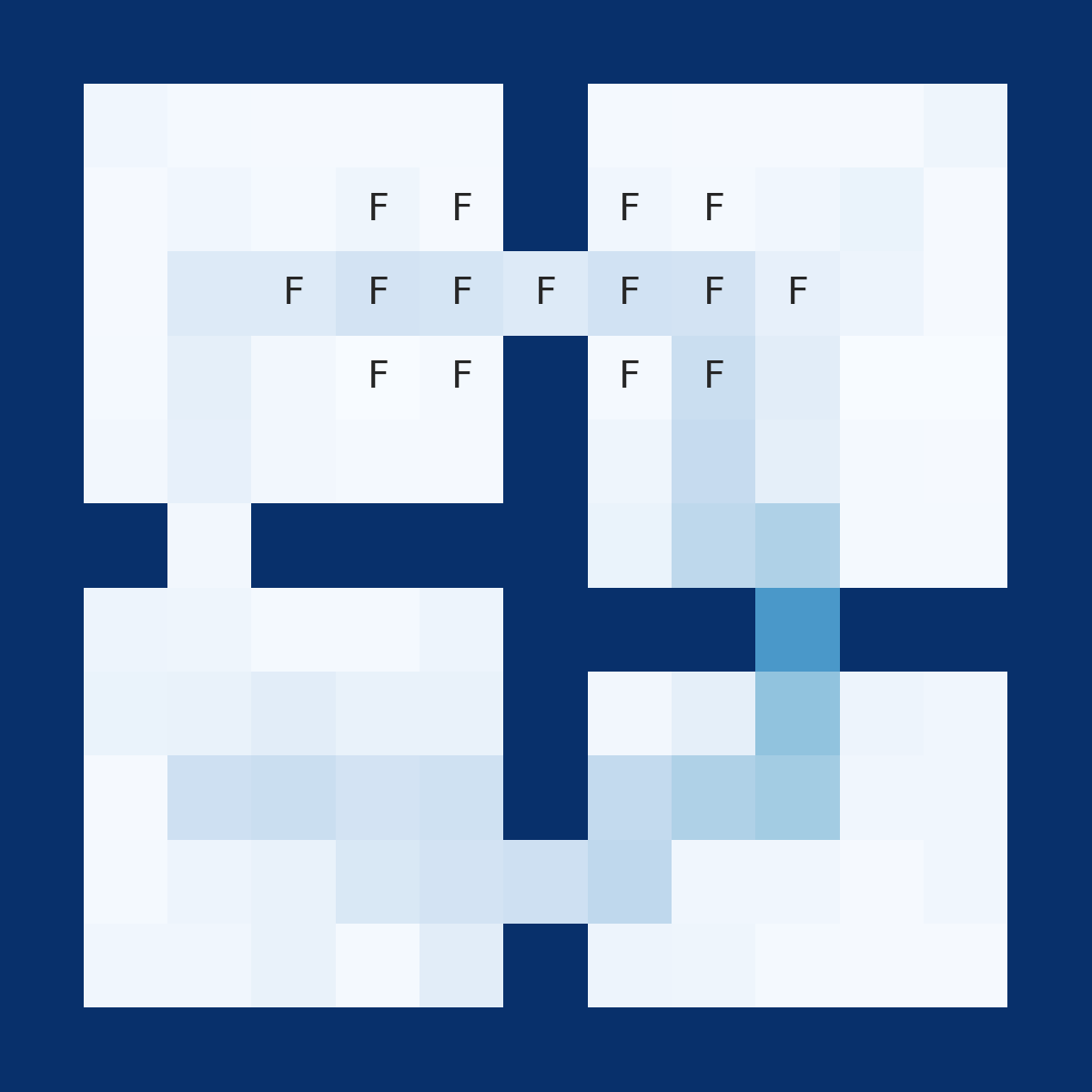}
			\caption[]%
			{{\small OC}}   
		\end{subfigure}
		\quad
		\begin{subfigure}[b]{0.25\textwidth}  
			\captionsetup{justification=centering}            \includegraphics[width=\textwidth]{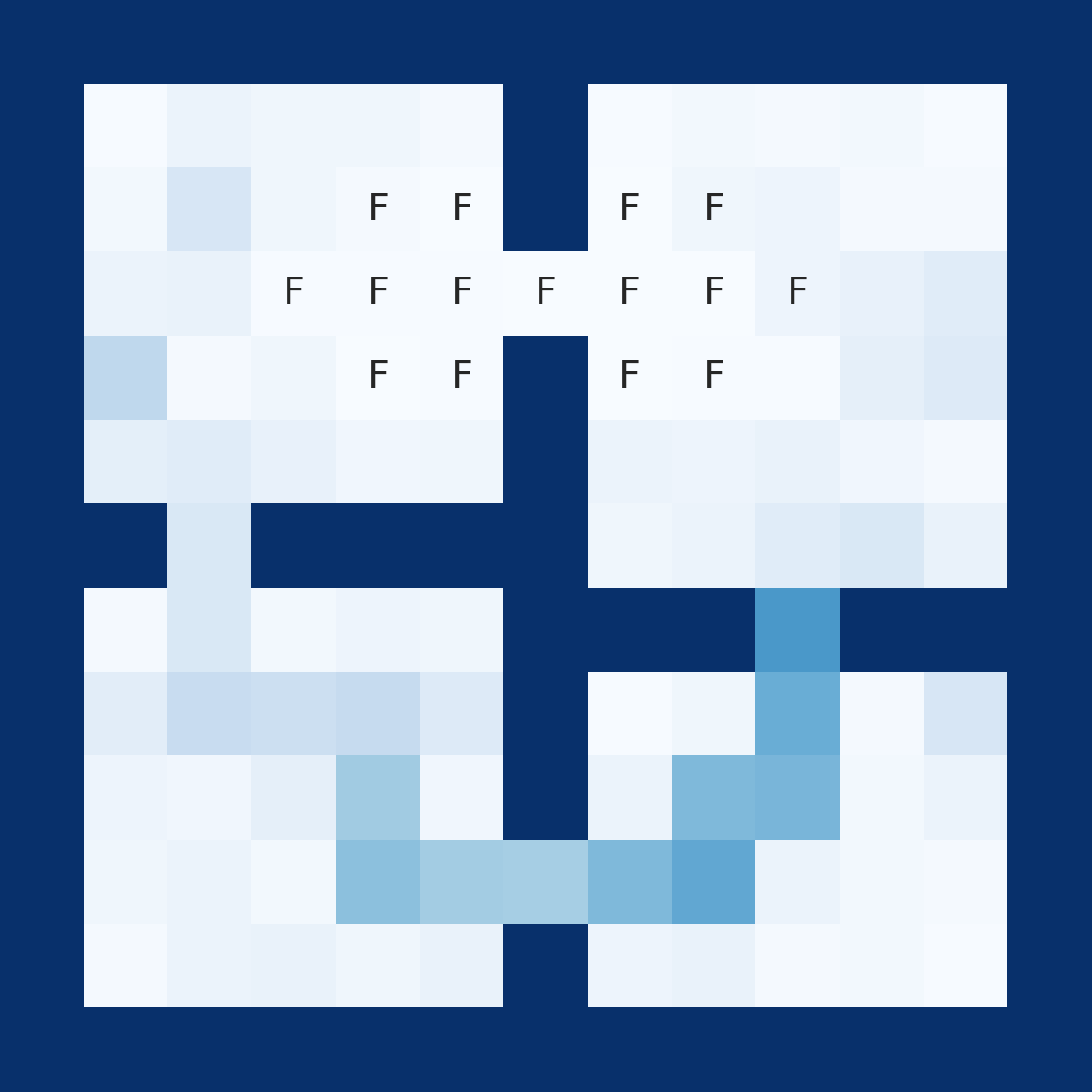}
			\caption[]%
			{{\small Safe-OC}}    
		\end{subfigure}
		\caption{\textbf{State visitation frequency in Four Room Environment.} Density graph represents the number of times a state was visited during testing over $80$ independent trials. Darker shades of blue represents a higher density. a) Model without safety has equally likely density for both the hallways. b) Model with safety shows higher density for the path without the frozen states.}
		\label{fig:Frequency_heat_map}
	\end{center} 
\end{figure}

It can be observed from Fig.~\ref{fig:FR_return} that the options with the controllability (Safe-OC) have a lower standard deviation in the return of an episode as compared to the baseline i.e. options without any notion of safety (OC). Fig.~\ref{fig:FR_TD_error} represents a lower standard deviation in the absolute sum of the TD error with safety. This highlights the fact that the controllability helps the agent to avoid the unsafe states (variable reward), thus resulting into  a smaller variance in the sum of absolute TD error over multiple trials. The Fig. \ref{fig:variance_regularizer_ fourroom} shows how the performance (expectation and variance in the return) varies with the change in the controllability regularizer $\psi$. We chose the optimal value of $\psi$ which maximizes the mean, but also, minimizes the variance in return. From the figure, it can be easily seen that $\psi=0.1$ leads to maximum expectation but also significant reduction in the variance of return. We also visualize the state frequency graph depicted in the Fig.~\ref{fig:Frequency_heat_map}. It is observed that the options with the controllability function have lower frequency of visit to the frozen states (risky) as opposed to the vanilla options. 

\begin{figure}[ht]
	\begin{center}
		\begin{subfigure}[ht]{\textwidth}
			\includegraphics[width=0.21\textwidth]{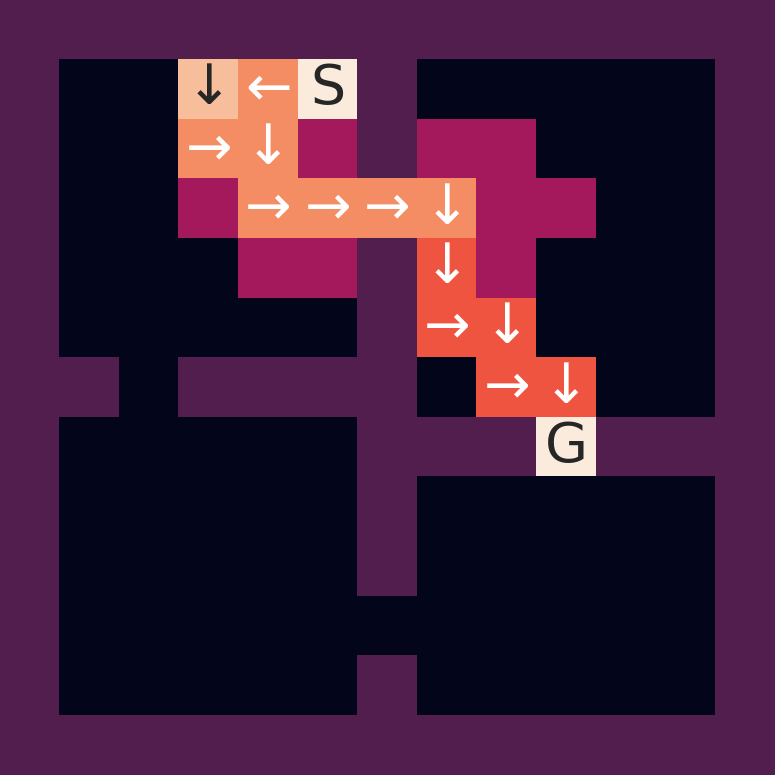}
			\hfill
			\includegraphics[width=0.21\textwidth]{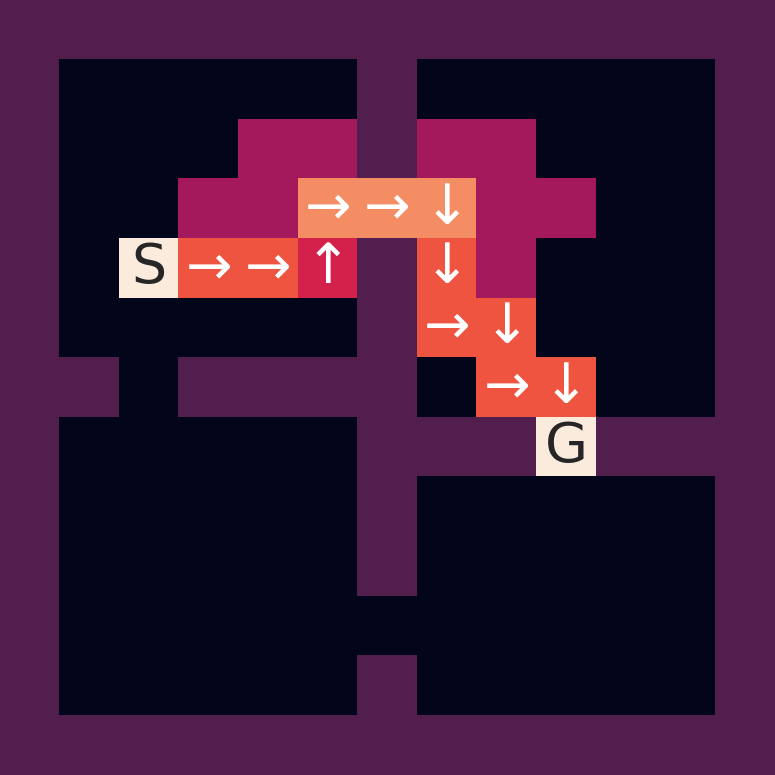}
			\hfill
			\includegraphics[width=0.21\textwidth]{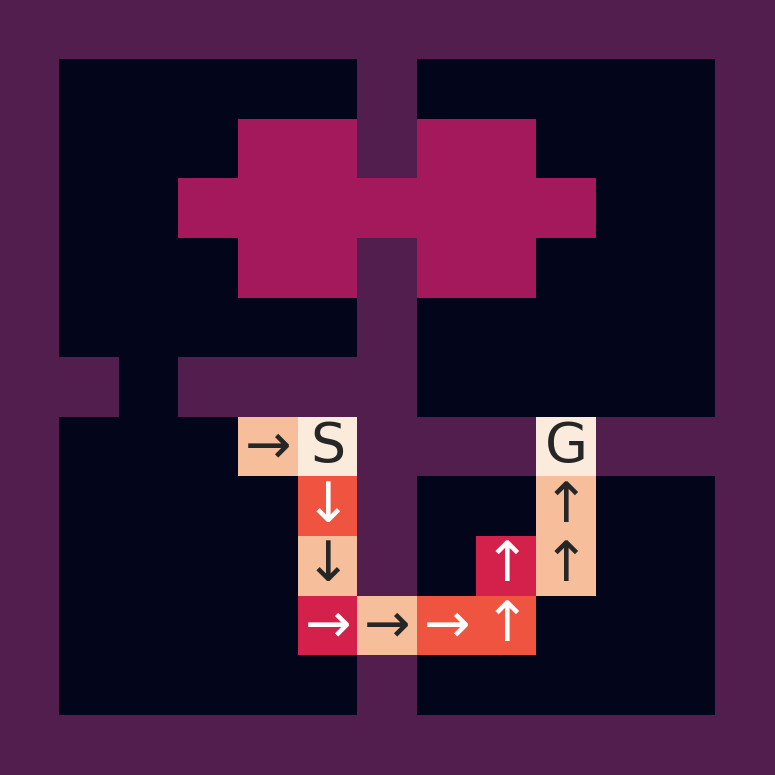}
			\hfill
			\includegraphics[width=0.21\textwidth]{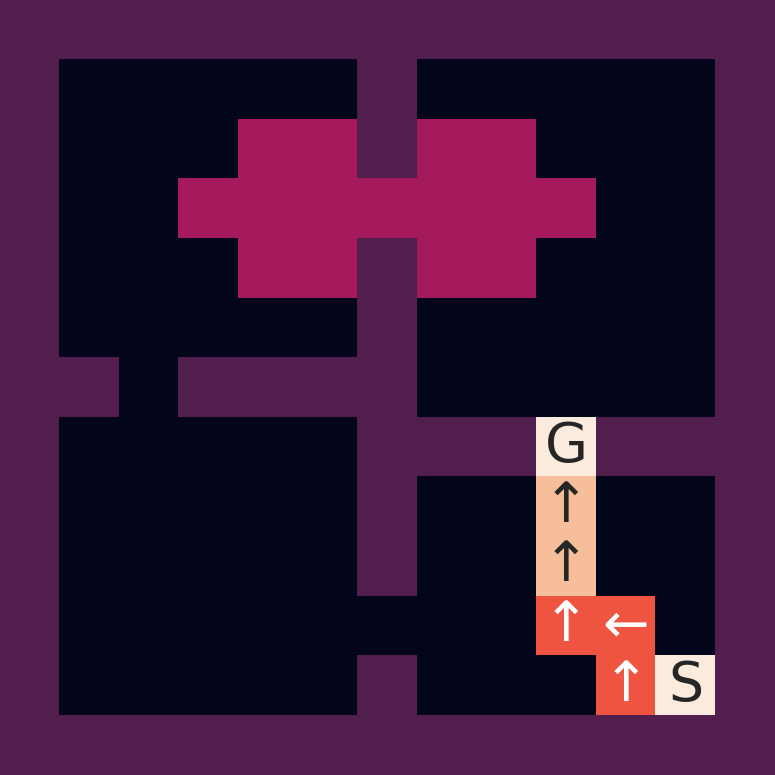}
			\caption[]%
			{{\small OC}} 
			\label{fig:OC_FR_Traj}
		\end{subfigure}
		\vskip\baselineskip
		\begin{subfigure}[b]{\textwidth}
			\includegraphics[width=0.21\textwidth]{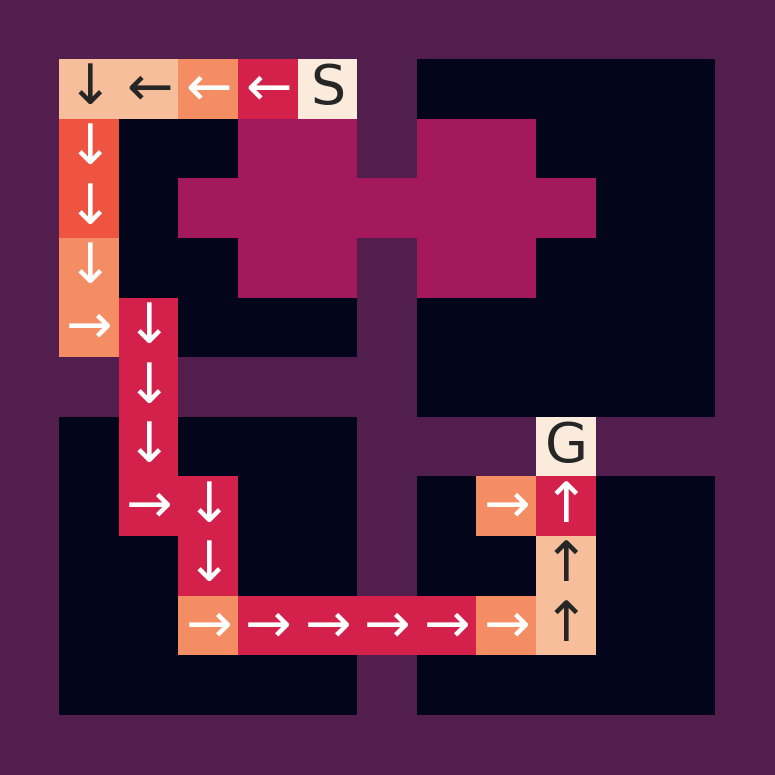}
			\hfill
			\includegraphics[width=0.21\textwidth]{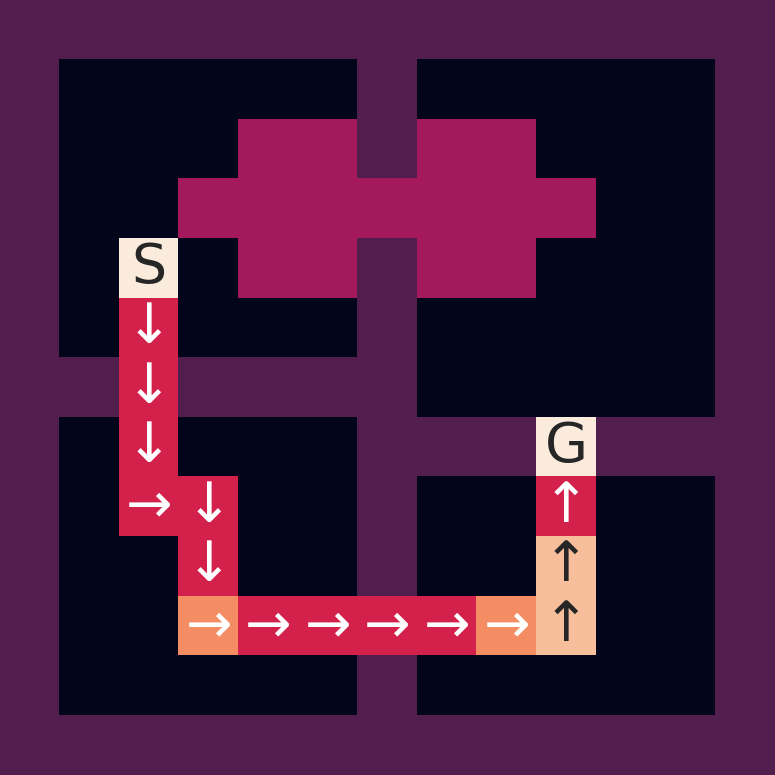}
			\hfill
			\includegraphics[width=0.21\textwidth]{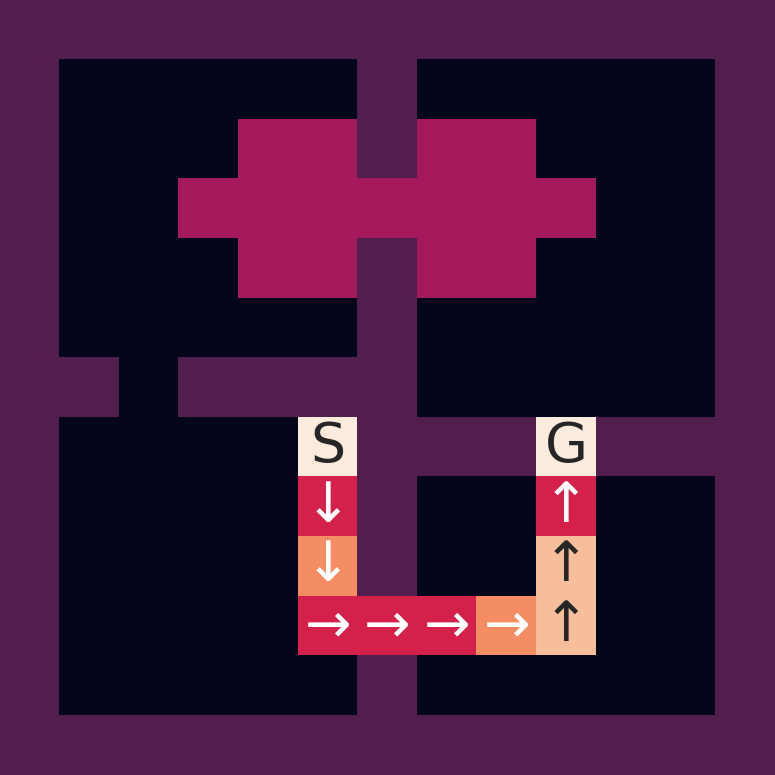}
			\hfill
			\includegraphics[width=0.21\textwidth]{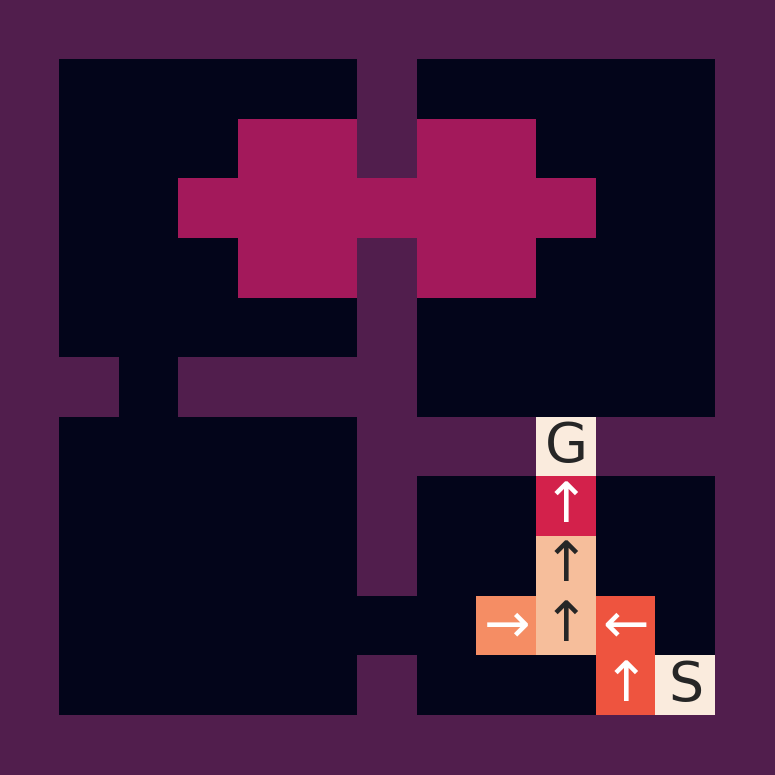}
			\caption[]%
			{{\small Safe-OC}} 
			\label{fig:S-OC_FR_Traj}
		\end{subfigure}
		\caption[]
		{\small \textbf{Policy in FR Environment.} Sampled
		trajectories learned with 4 options where $S$ and $G$ represents the start \& goal state. Arrows denote the 4 actions. Agent might take different actions due to environment stochasticity. Change in color represents the option switching. Same color represents the same option choice. Purple patch represents the frozen states. a) depicts the policy with $\psi=0$ passing through the frozen area. b) depicts the policy learned with $\psi=0.1$ that avoids the frozen area due to the inbuilt safety constraint.} 
		\label{fig:policy_fourroom}
	\end{center}
\end{figure}

Learning of safe options induces transparency in the behaviour of an agent. This is most explicitly demonstrated through the trajectory taken by the agent without controllability regularizer \ (baseline OC) and with controllability regularizer (Safe-OC) as shown in the Fig.~\ref{fig:policy_fourroom}. Regardless of the start state, Safe-OC agent navigates to the goal state by avoiding the states with a high variance in the reward, as opposed to the OC agent, which finds a shortest route to the goal state being unaware of the risky states.

\begin{figure}[ht]
		\begin{subfigure}[b]{\textwidth}
			\includegraphics[width=0.19\textwidth]{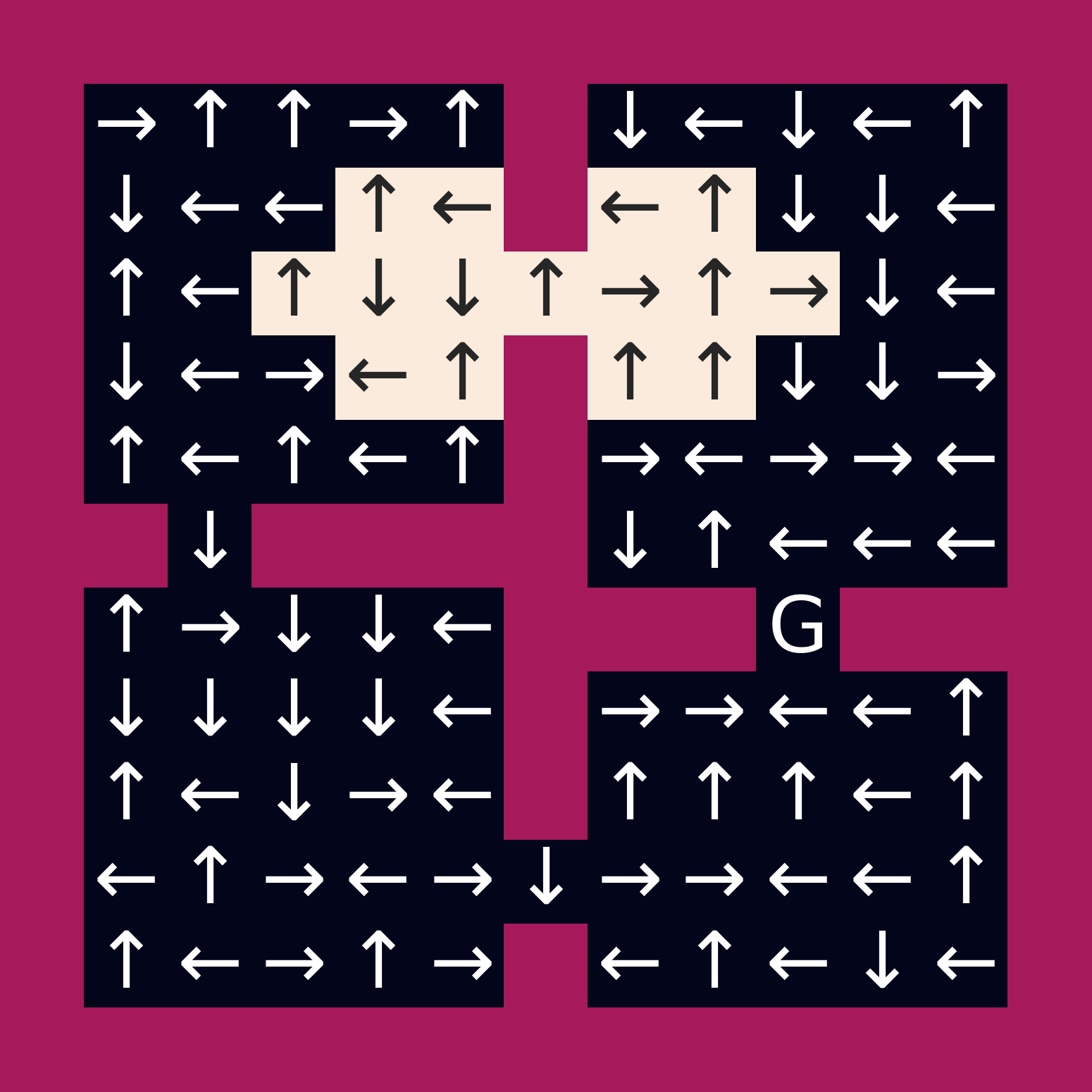}
			\includegraphics[width=0.19\textwidth]{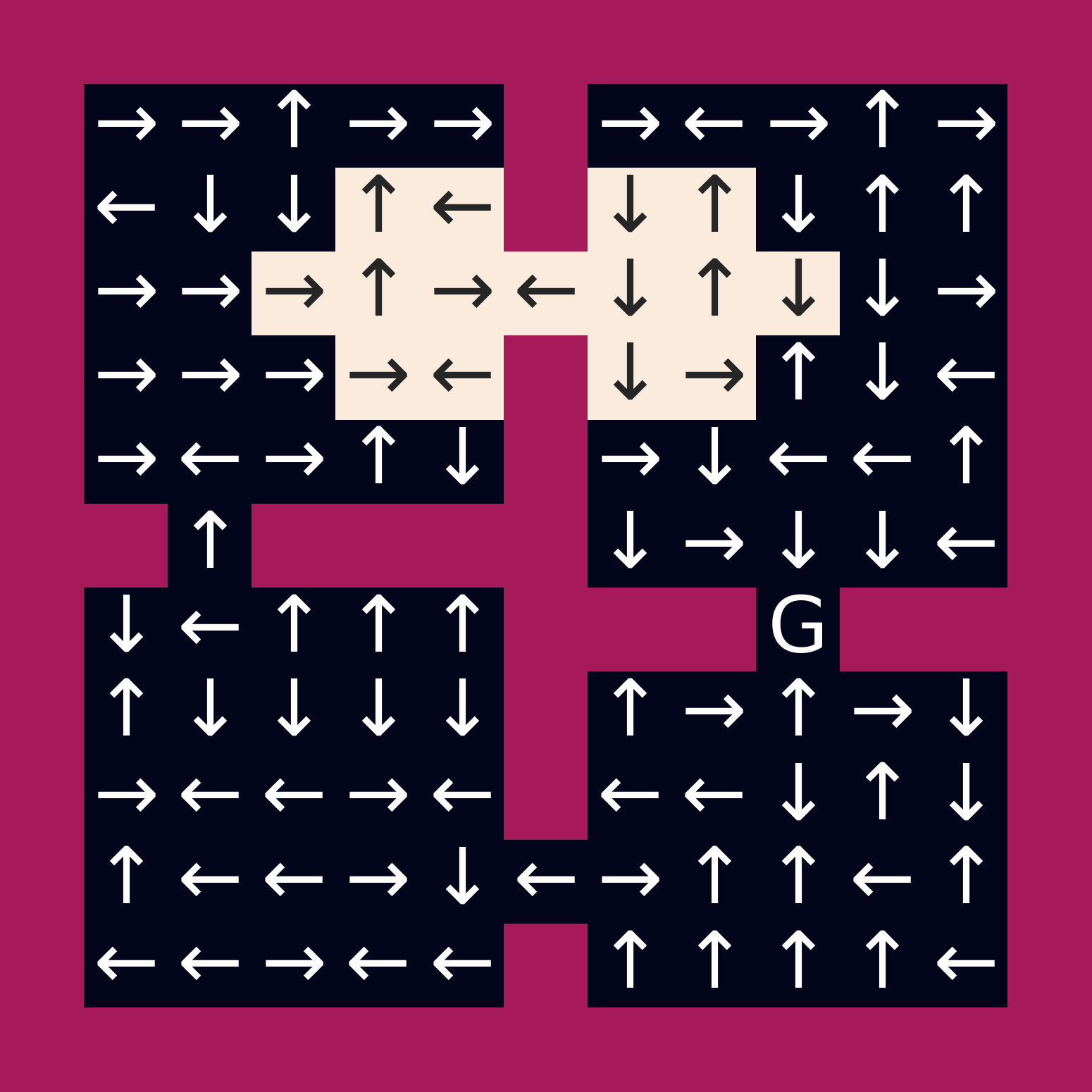}
			\includegraphics[width=0.19\textwidth]{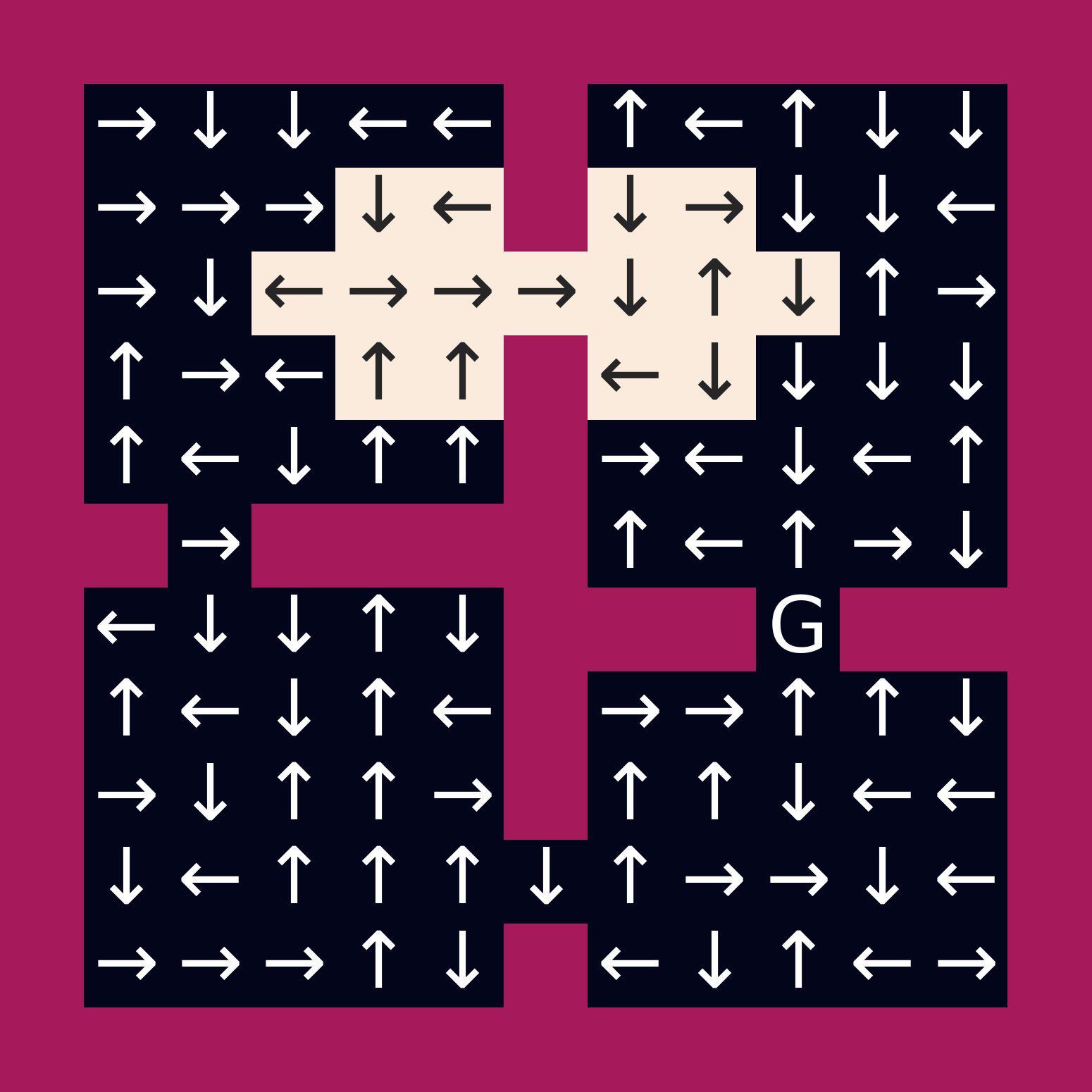}
			\includegraphics[width=0.19\textwidth]{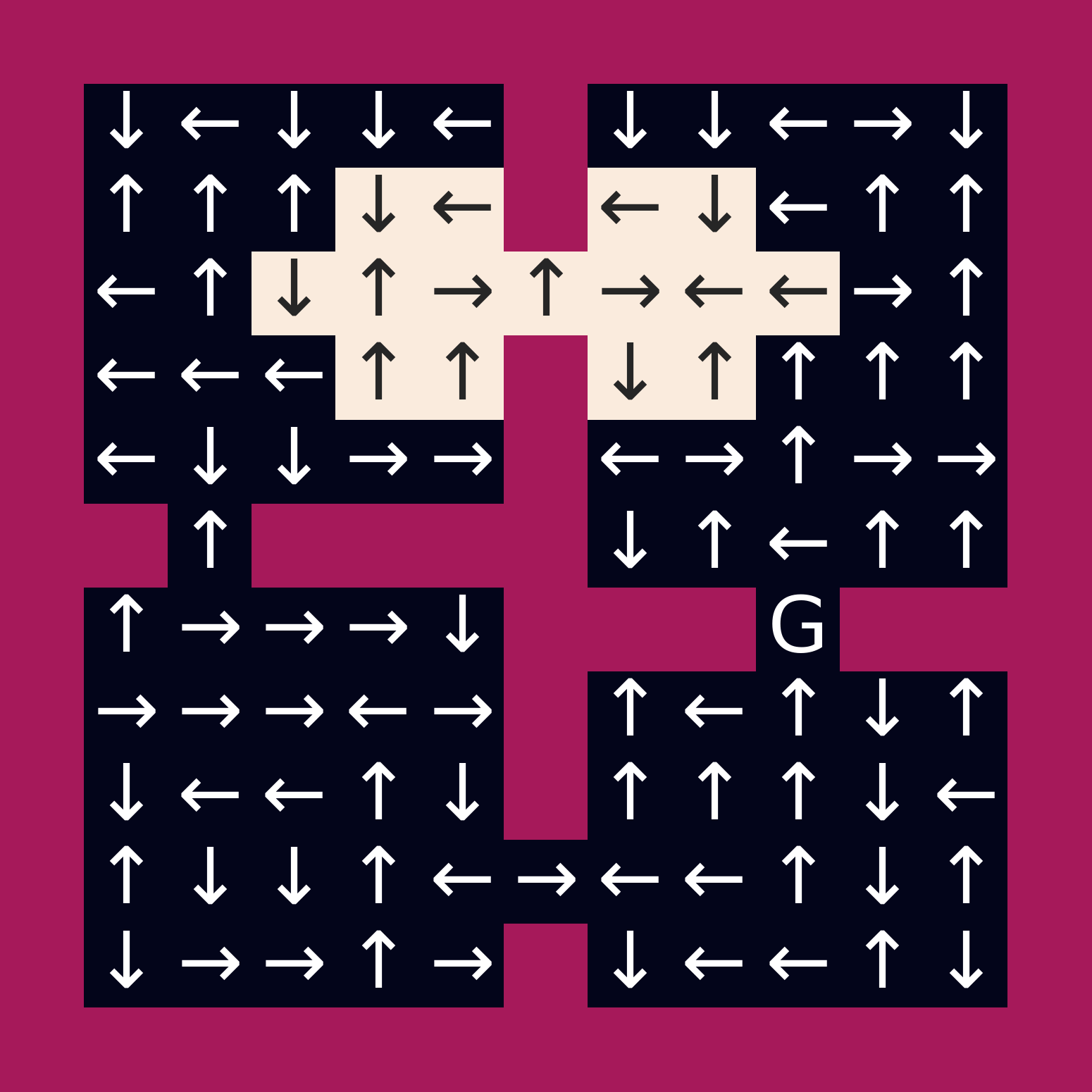}
			\includegraphics[width=0.19\textwidth]{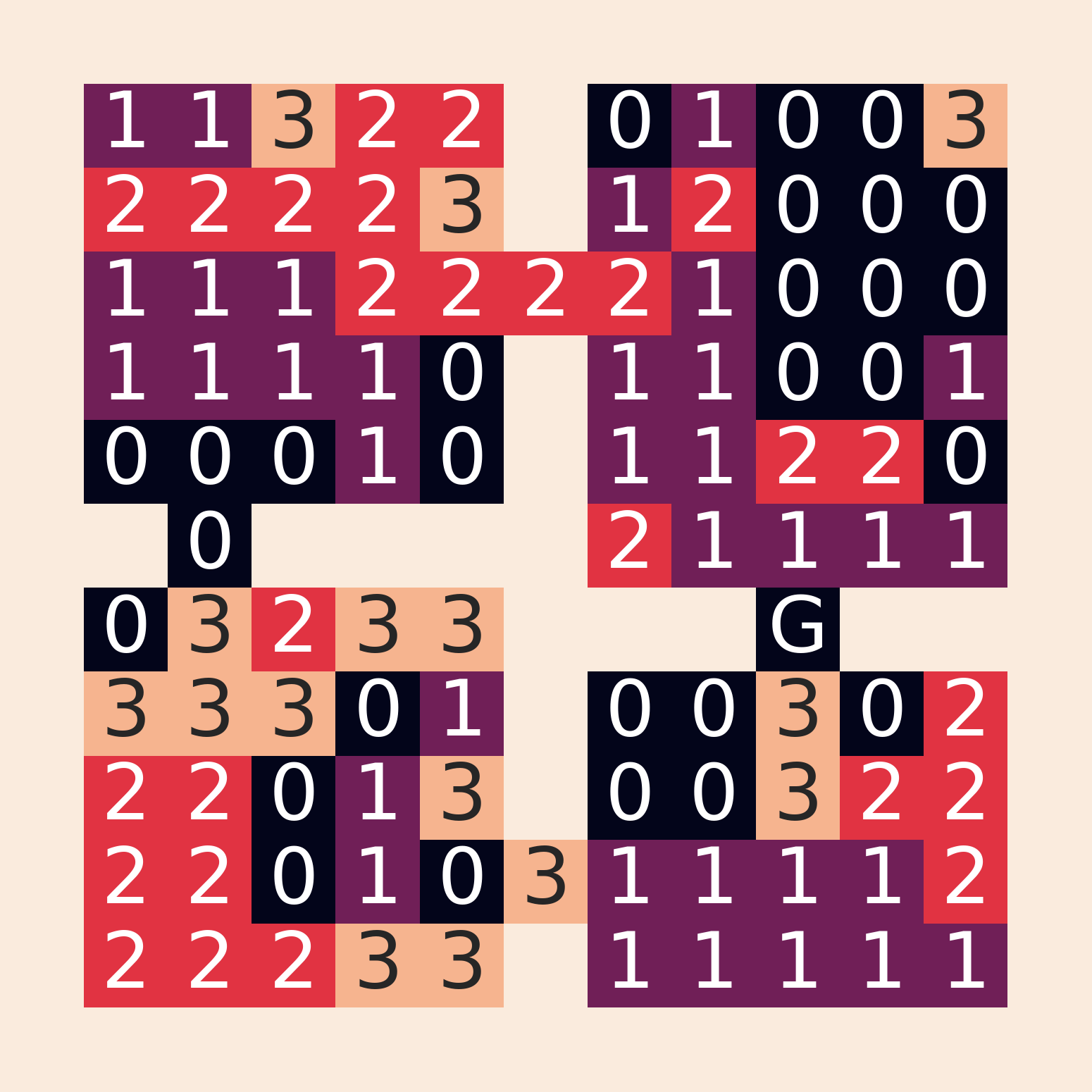}
			\caption[]%
			{{\small Baseline OC Policy in 4 Options}}    
			\label{fig:S_P}
		\end{subfigure}
		\begin{subfigure}[b]{\textwidth}
			\includegraphics[width=0.19\textwidth]{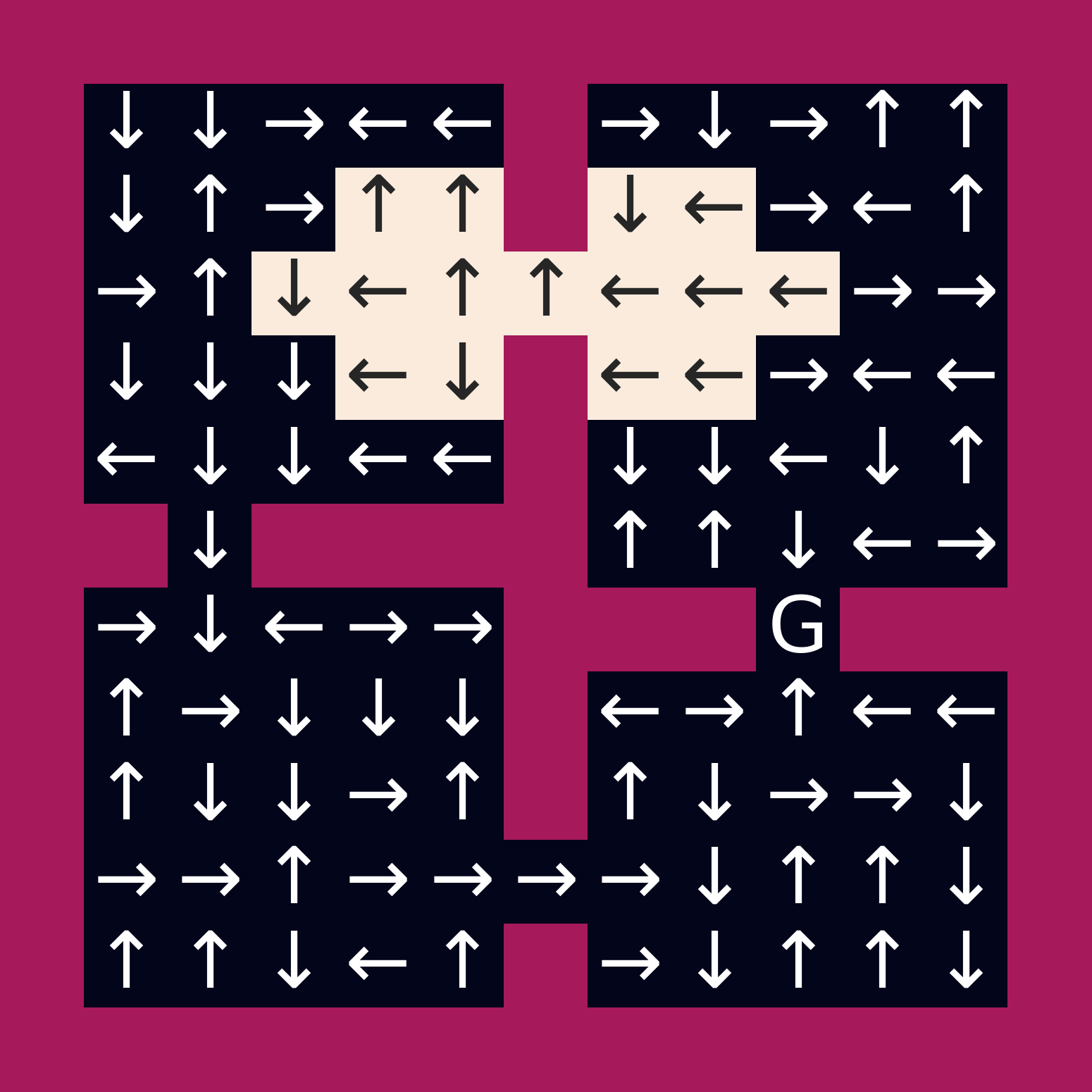}
			\includegraphics[width=0.19\textwidth]{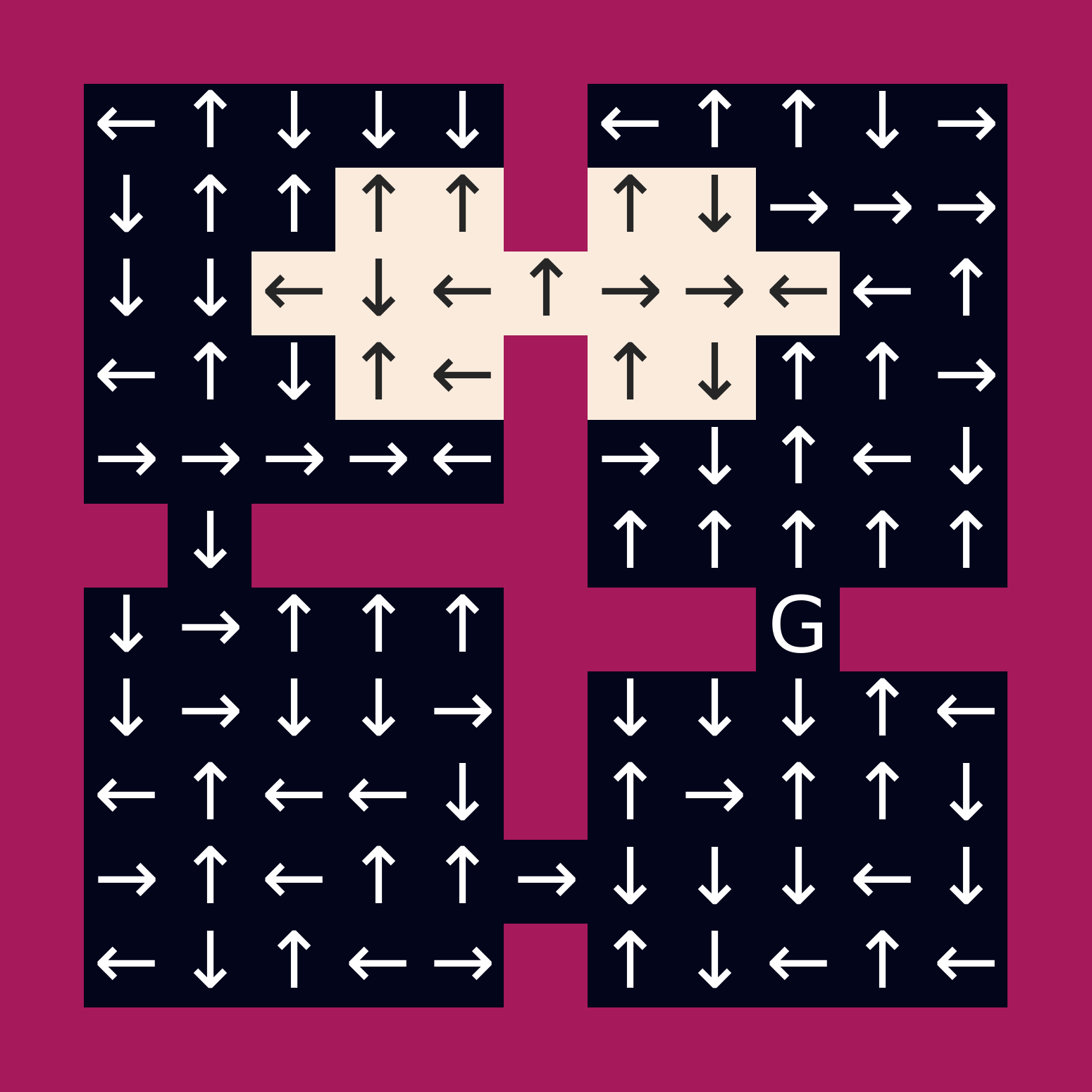}
			\includegraphics[width=0.19\textwidth]{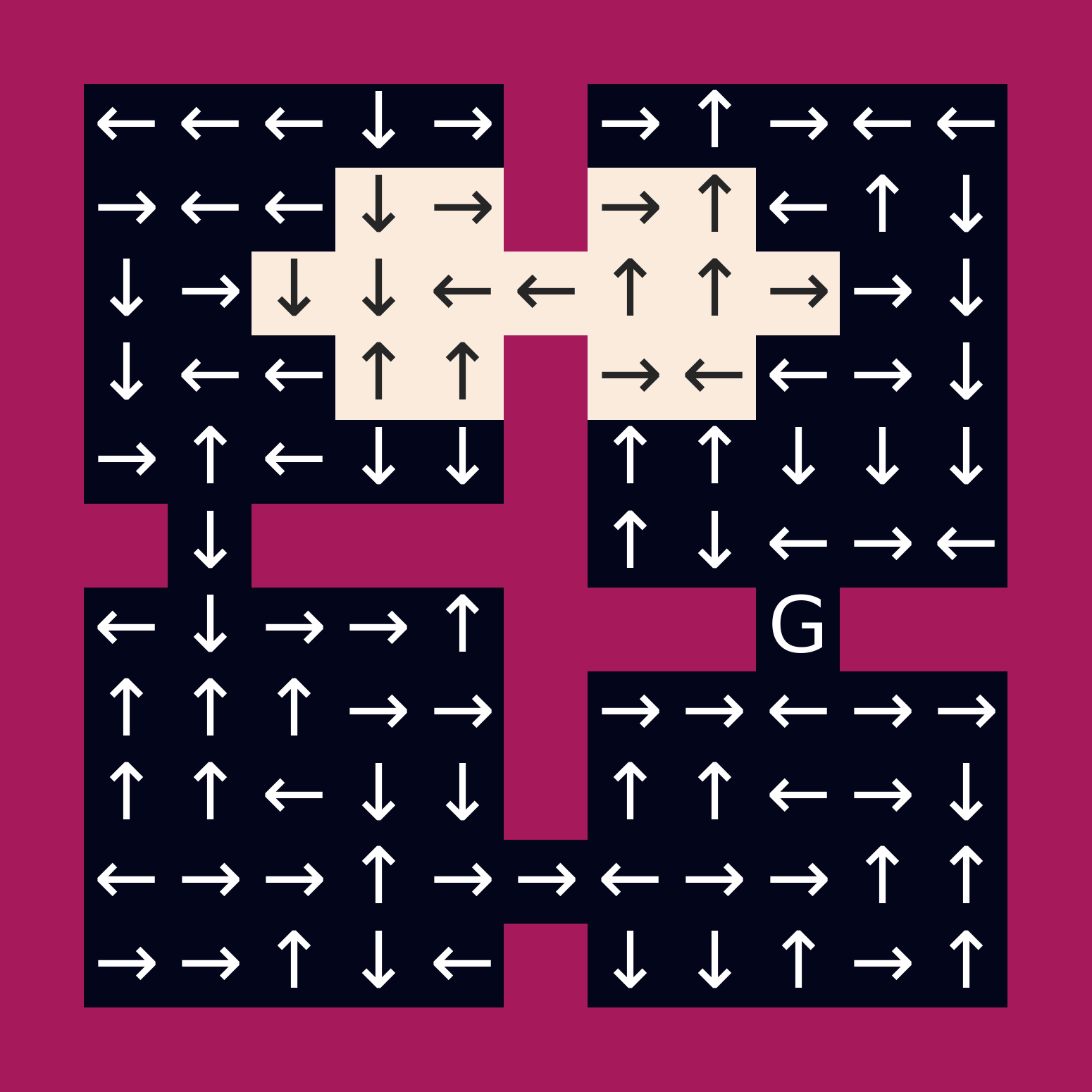}
			\includegraphics[width=0.19\textwidth]{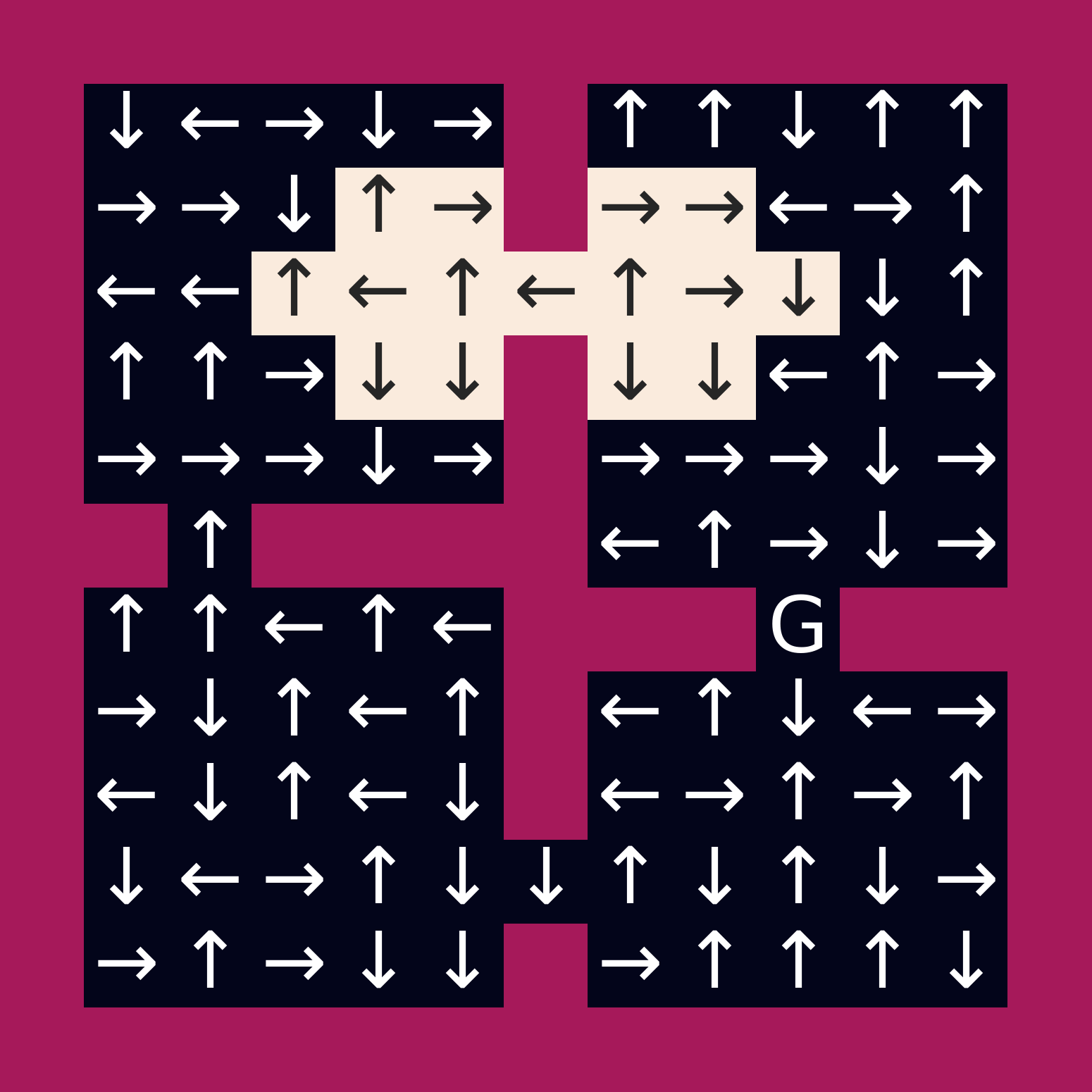}
			\includegraphics[width=0.19\textwidth]{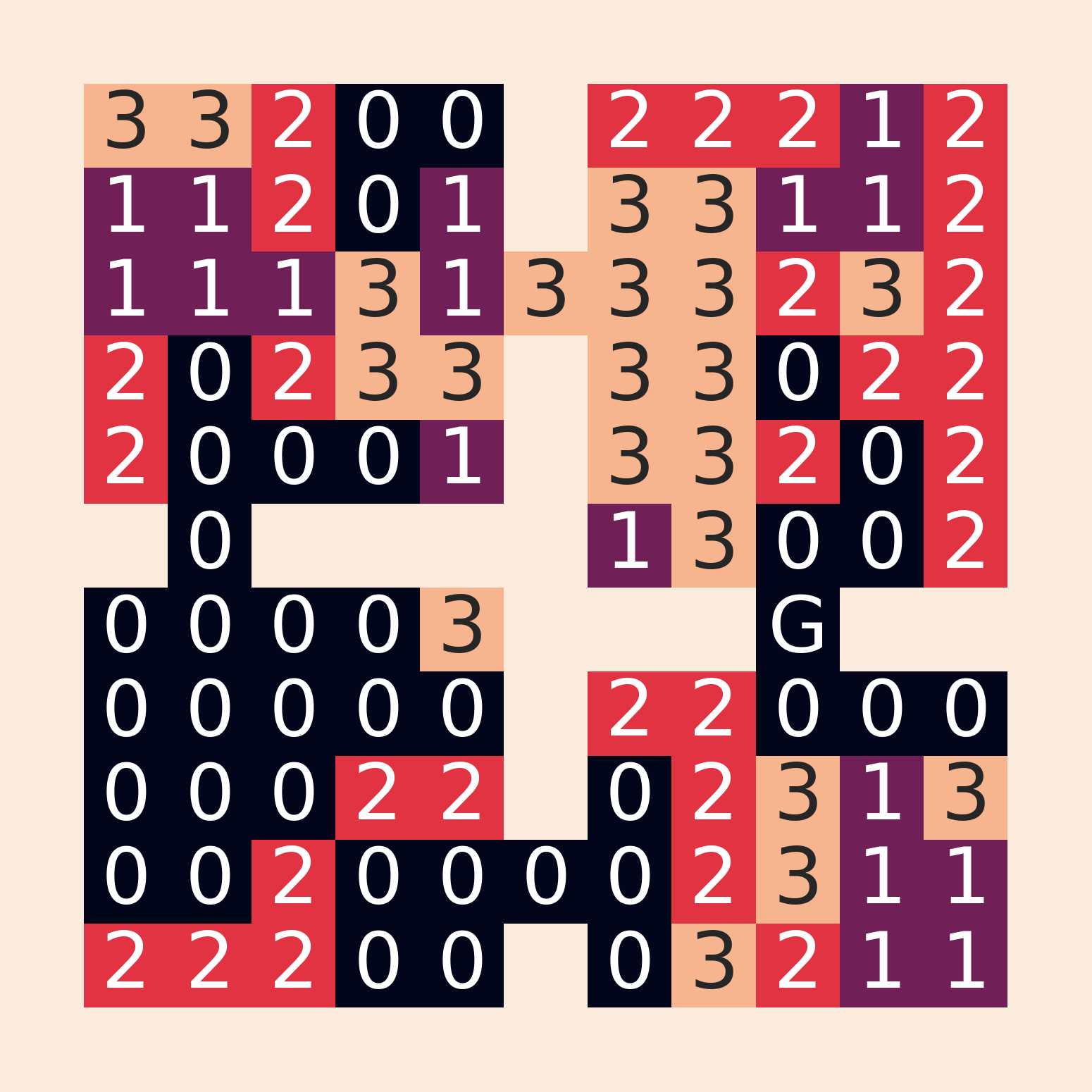}
			\caption[]%
			{{\small Safe OC Policy in 4 Options}}    
		\end{subfigure}
		\caption[]
        {\textbf{Converged policy for $4$ options in fourrooms environment.} The first four figures in each row depicts the intra-option policy for all the four options. The last figure in both the rows show the policy over the options where the number depicts the different options(numbering of options increases from left to right image in a row). Both the baseline and the Safe OC exhibits that the overall hierarchical policy is more reasonable than the individual option policy. Though the overall trend of an individual safe option policy is to avoid the frozen hallway.} 
        \label{fig:ConvPol_4option}
\end{figure}

To capture how each option is behaving, we plotted the converged policy for all the options in Fig. \ref{fig:ConvPol_4option}. It can be observed that each option does not capture the safety for the entire state space, but rather a subset of state space. The switches among the options for the safety are similar to the baseline (last image in both the rows). The overall trend observed from the individual option policy of safe algorithm is to avoid the unsafe frozen patch whereas, the vanilla option policies do not differentiate between the two hallways.

\subsection{Continuous State Space Environment}
In this section, we examine the performance in the puddle-world Open AI gym environment with the linear function approximation. It is a continuous 2-dimensional state-space  environment in $[0,1]^2$. The action space consists of four actions that change the position by a value of $0.05$ in any one of the coordinates. In each transition, noise is drawn from the $\textit{Uniform}[-0.025, 0.025]$ distribution, which is added to both the coordinates. For introducing the notion of unsafe regions, similar to the four rooms environment, we added a square puddle region with a center at $(0.5, 0.5)$ and $0.4$ height. An agent can be randomly initialized from any state and the goal is located at $(1,1)$. When the agent is within $0.1$ L$1$ distance from the goal state, the agent receives a reward of $50$. Whenever the agent transitions from a puddle region, it receives a reward from a normal distribution of $\mathcal{N}(\mu=0, \sigma=8)$,  and  $0$ reward otherwise. On expectation, the reward received from both regions is the same. The penalty makes the puddle locations `unsafe' due to the variability in the rewards.

For these experiments, we use $2$ options. One could choose to have more number of options, but given the complexity of the task, we decide to use only $2$ options.  We use intra-option Q-learning in the critic to learn policy over options. Boltzmann distribution is used to represent both intra-option policies and  policy over options with a temperature value of $0.1$. To learn the termination function, we use sigmoid activation. The agent can take a maximum of $5000$ steps in an episode. The standard tile coding \citep{Sutton:1998:IRL:551283} is used for discretizing the state-space. We use $5$ tilings, each of $5 \times 5$ over the joint space of two features which is hashed to a vector of $1024$ dimension. The discount factor $\gamma$ is set to $0.99$. The optimized parameters are shown in the Table~\ref{tab:param-OC}. The code for both the discrete four-rooms and the continuous puddle-world is available at Github\footnote{\textbf{Code} available on \url{ https://github.com/arushi12130/SafeOptionCritic}.}.

\begin{figure}[ht]
	\begin{center}
		\centerline{\includegraphics[width=0.6\columnwidth]{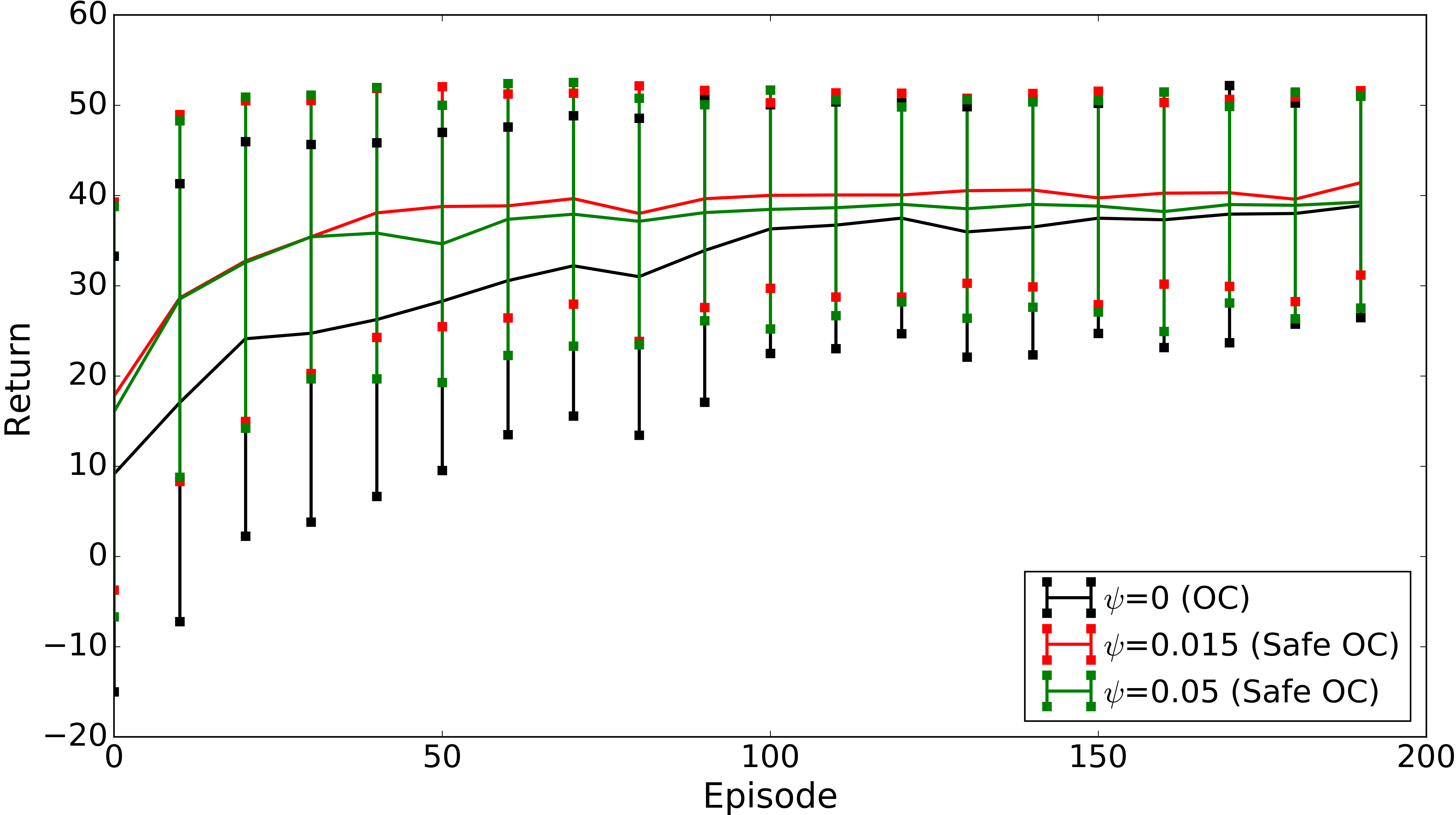}}
		\caption{\textbf{Puddle-World Environment}: The graphs depicts the return averaged over $50$ independent trials with $2$ options. The vertical bands show the standard deviation in the return over multiple trials. The experiment with safe policy $\psi=0.015$ (red) has a smaller standard deviation as compared to the baseline (black) signifying safety helps an agent to avoid variance inducing regions.}
		\label{fig:puddle}
	\end{center}
	\vskip -0.2in
\end{figure}

\begin{figure}[h]
	\begin{center}
		\begin{subfigure}[b]{0.22\textwidth}
			\centering
			\captionsetup{justification=centering}
			\includegraphics[width=\textwidth]{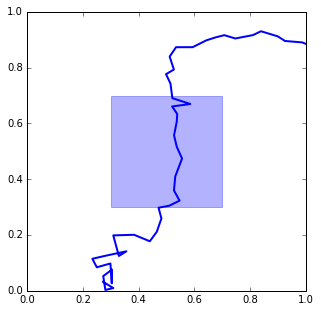}
			\caption[]%
			{{\small OC}}    
			\label{fig:mean and std of net14}
		\end{subfigure}
		\quad
		\begin{subfigure}[b]{0.22\textwidth}  
			\centering
			\captionsetup{justification=centering}
			\includegraphics[width=\textwidth]{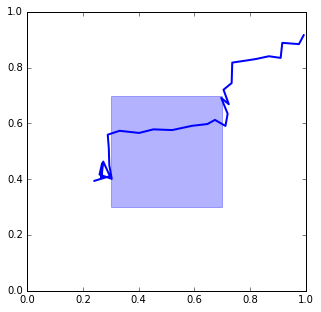}
			\caption[]%
			{{\small OC}}    
			\label{fig:mean and std of net24}
		\end{subfigure}
		\quad
		\begin{subfigure}[b]{0.22\textwidth}   
			\centering
			\captionsetup{justification=centering}
			\includegraphics[width=\textwidth]{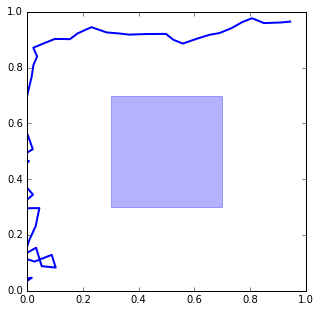}
			\caption[]%
			{{\small Safe-OC}}    
			\label{fig:mean and std of net34}
		\end{subfigure}
		\quad
		\begin{subfigure}[b]{0.22\textwidth}   
			\centering 
			\captionsetup{justification=centering}
			\includegraphics[width=\textwidth]{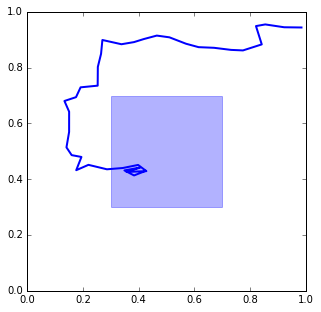}
			\caption[]%
			{{\small Safe-OC}}
		\end{subfigure}
		\caption[]
		{\small \textbf{Trajectories in Puddle-World}: The graph shows sampled trajectories from OC policy in (a) \& b, and Safe-OC policy in c) \& d. With the designed safety constraint, the agent learns to bypass the unsafe central puddle.} 
		\label{fig:traj_PW}
	\end{center}
\end{figure}

\begin{figure}[ht]
	\begin{center}
		\begin{subfigure}[b]{0.35\textwidth}
			\centering
			\captionsetup{justification=centering}
			\includegraphics[width=0.84\textwidth]{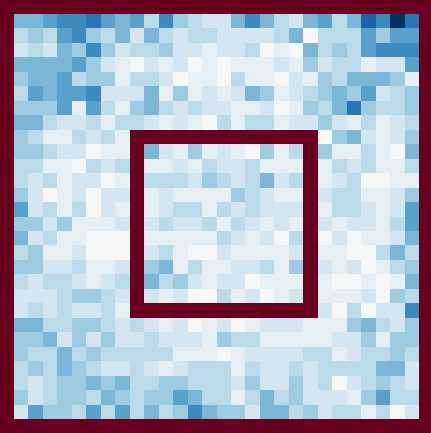}
			\caption[]%
			{{\small OC}}    
			\label{fig:F_U}
		\end{subfigure}
		\quad
		\begin{subfigure}[b]{0.35\textwidth}  
			\centering
			\captionsetup{justification=centering}
			\includegraphics[width=\textwidth]{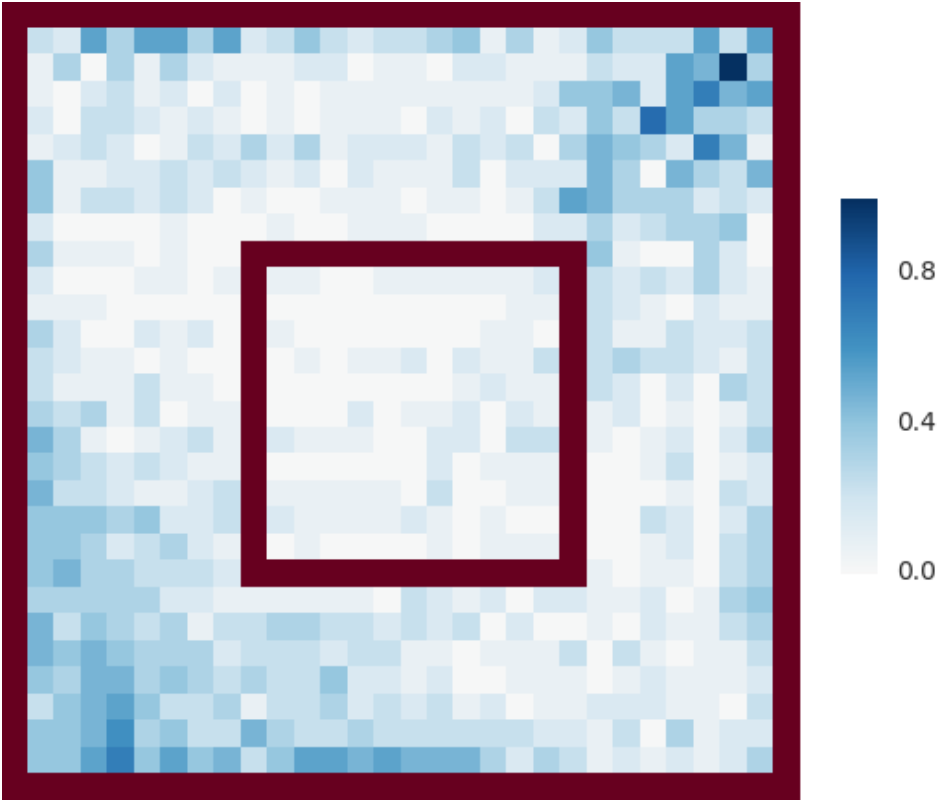}
			\caption[]%
			{{\small Safe-OC}}    
			\label{fig:F_S}
		\end{subfigure}
		\caption[]
		{\small \textbf{Visitation Frequency in Puddle-World Environment}: depicts the state visitation frequency averaged over $50$ independent trials for a particular episode in the a) baseline OC and (b) Safe-OC. The graph highlights that the Safe-OC agent's visit to the unsafe central puddle region decreases in comparison to the OC agent. Our notion of safety enforces the agent to avoid the puddle region.
		} 
		\label{fig:freq_TD_puddle}
	\end{center}
\end{figure}

Fig.~\ref{fig:puddle} shows the return observed in the puddle-world domain with different values of $\psi$ averaged over $50$ trials. The best performance in terms of stability was observed by $\psi=0.015$ Safe-OC which exhibits a reduction in the standard deviation of the return across multiple runs as compared to the baseline ($\psi=0$). This highlights that the agent following the safe policy learns to avoid the variance inducing puddle region. Fig.~\ref{fig:traj_PW} shows the sampled trajectories from both the baseline OC and the Safe-OC. To further validate, the frequency of visits is shown in  Fig.~\ref{fig:F_U} \& Fig.~\ref{fig:F_S}. They describe the average number of visits across multiple runs for an episode, depicting a decrease in the number of visits to the unsafe puddle region in the safe policy learning option-critic framework as compared to the baseline. It  demonstrates that the safe policy learning agent avoids the erratically behaving unsafe puddle region placed at the centre of the state space.

\begin{table}[h]
\centering
\caption{\textbf{Parameters for discrete and continuous  environment}: Optimized parameters for both four rooms and puddle-world environments}
\label{tab:param-OC}
\begin{tabular}{l|llllll}
\hline
\textbf{Type} &\boldmath{$\psi$} & \boldmath{$\alpha_c$} & \boldmath{$\alpha_\theta$} & \boldmath{$\alpha_\nu$} & \boldmath{temp} & \boldmath{option} \\ \hline
\textbf{Four}   & \textbf{0.0}     & 0.1    & 0.01  & 0.01 & 0.001 & 4        \\
\textbf{Room}   & \textbf{0.1}   & 0.5    & 0.01  & 0.1  & 0.001 & 4        \\\hline
\textbf{Puddle} & \textbf{0.0}     & 0.1    & 0.01  & 0.01 & 0.1   & 2        \\
\textbf{World}  & \textbf{0.015} & 0.5    & 0.05  & 0.05 & 0.1   & 2       \\\hline
\end{tabular}
\end{table}

\subsection{Arcade Learning Environment}
We now analyze our method in the ALE domain~\citep{bellemare2013arcade}. \cite{harb2017waiting} introduced deliberation cost in the options framework to learn temporally extended options by adding a penalty cost on switching options too frequently. We use the asynchronous advantage option-critic (A2OC)~\citep{harb2017waiting} as our baseline for learning `safe' options with non-linear function approximation. A2OC provides an extension of the asynchronous advantage actor-critic (A3C) algorithm~\citep{mnih2016asynchronous} in the option-critic architecture. Introducing the controllability function in the A2OC algorithm results in an additional regularizer term in the intra-option policy gradient (see Equation \eqref{j_grad_theta}). The update rule for the intra-option policy gradient in the A2OC with controllability is given as follows:
\begin{equation}
	\begin{split}
		\theta_{\pi} \leftarrow\ & \theta_{\pi} + \alpha_{\theta_\pi} \Big \{\frac{\partial \log \pi_{w, \theta}(a|s)}{\partial \theta}\big( G - Q_\Theta(s, w) \big)
		- \underbrace{\psi \frac{\partial \log\pi_{w_0, \theta}(a_0|s_0)}{\partial \theta} \delta^2(s_0,w_0,a_0)}_{\text{controllability}} \Big \}.
	\end{split}
\end{equation}
Here, similar to the A2OC algorithm, $G$ is a mixture of $n$-step return, with a slight modification that the $n$-step return is only considered until the current option terminates or number of steps are more than $n$. Without any loss in generality, the $1$-step TD error in the definition of controllability can be extended to the $n$-step TD error if the current option persists until the $n^{th}$ step. Similarly, as discussed in the Equation \eqref{eq_J_v_f}, there is no change in the termination gradient and we use the same update rule as derived in the A2OC algorithm. In the below equation, $\eta$ represents the deliberation cost \citep{harb2017waiting} which penalizes the frequent switching of the options, 
\begin{equation}
	\nu \leftarrow\ \nu - \alpha_\nu \frac{\partial \beta_{w, \nu}(s')}{\partial \nu}\big( Q_{\Theta}(s', w) - V_W(s') + \eta \big).
\end{equation}
We evaluate the performance in three games, namely, MsPacman, Amidar, and Q*Bert from the ATARI $2600$ suite. We introduce Safe-A2OC \footnote{\textbf{Code} is available at \url{https://github.com/kkhetarpal/safe_a2oc_delib}.} which uses a similar deep network architecture as A2OC but with an additional safety criteria added to it. We use $\epsilon$-greedy for learning policy over options. Intra-option policies follows the linear Boltzmann distribution. Termination functions use sigmoid activation along will linear function approximation for the Q values. For hyperparameters, we learn $4$ options, with a fixed deliberation cost of $0.02$, margin cost of $0.99$, step size of $0.0007$, and entropy regularization of $0.01$ for varying degrees of controllability ($\psi$) and $\epsilon$. The training used $16$ parallel threads for all our experiments. We also optimize for $\epsilon$ parameter for the baseline case ($\psi=0$). For fair analysis, we compare the best performance of A2OC with varying $\psi$ regularizer for controllability function in the Safe-A2OC. 

\textbf{Results and Evaluation}: 
To evaluate the agents, we investigate the averaged performance over $100$ games after the training has been completed over $80$M frames~\citep{MarlosEvaluationProtocol}. Table \ref{tab:finalscores} depicts the averaged performance over $5$ different runs, where, each run contains the averaged score obtained over 100 testing games. It is observed that the Safe-A2OC with a controllability value of $\psi = 0.10$ in MsPacman not only outperforms the score achieved by the baseline  A2OC ($\psi= 0$), but also, results in a smaller variance in the scores (values are shown in the braces). We observe a similar pattern in Q*Bert, wherein, the Safe-A2OC ($\psi = 0.05$) shows an improved robust performance over the A2OC ($\psi=0$), achieving almost half of the variance of A2OC across multiple games.
\begin{table*}[ht]
	\centering
	\caption{ \textbf{ALE Final Scores}: The performance is averaged over $100$ games once the training is completed. The scores are averaged over $5$ independent seeds. Scores in the boxes highlight the performance without controllability function (baseline), whereas, aqua highlighted cells indicate the benefits of introducing our notion of safety in learning end-to-end options. \textbf{Introducing controllability in options outperforms best performances of primitive actions (gray) in $2$ out of $3$ games analyzed here.} \textbf{Learning options with our notion of safety outperforms the baseline A2OC in all $3$ games}. A3C scores have been taken from \cite{mnih2016asynchronous}, DQN from \cite{Nair}, Double DQN from \cite{van2016deep}, and Dueling from \cite{WangFL15}. $\psi$ represents the degree of controllability. Values in the brackets indicate the standard deviation across $100$ games for $5$ independent runs. \label{tab:finalscores}} 
	
	\begin{tabular}[b]{||c || c|| c|| c||} 
		\hline
		Algorithm & MsPacman & Amidar & Q*Bert\\ 
		\hline\hline
		A3C & $850.7$ & $\cellcolor{lightgray}283.9$ & $\cellcolor{lightgray}21307.5$\\ 
		DQN & $763.5$ & $133.4$ & $4589.8$\\ 
		Double DQN & $1241.3$ & $169.1$ & $11020.8$\\ 
		Dueling & $\cellcolor{lightgray}2250.6$ & $172.7$ & $14175.8$\\ 
		\hline\hline
		$\psi = 0$, $\epsilon = 0.2$ & \boxed{$2285.4 (756.64)$} & \boxed{$760.71 (204.08)$} & \boxed{$16881.25 (6107.04)$}\\ 
		$\psi = 0.05$, $\epsilon = 0.2/0.3$ & $2481.2  (909.48)$ & $569 (158.77)$ & $\cellcolor{aqua}\textbf{17642.0 (3346.85)}$\\
		$\psi = 0.10$, $\epsilon = 0.2$ & $\cellcolor{aqua}\textbf{2710.9 (598.69)}$ & $\cellcolor{aqua}\textbf{925.43 (211.52)}$ & $14490.0 (5962)$\\
		$\psi = 0.15$, $\epsilon = 0.2$ & $2055.8 (468.09)$ & $781.31 (168.79)$ & $1477.5 (961.85)$\\
		$\psi = 0.25$, $\epsilon = 0.2$ & $2290.4 (855.00)$ & $458.82 (107.77)$ & $298.25 (133.71)$\\
		\hline\hline
	\end{tabular}
\end{table*}

In the game of Amidar, we note that Safe-A2OC outperforms the score achieved by A2OC, and the other state-of-the-art approaches \citep{mnih2016asynchronous, Nair, van2016deep,  WangFL15} using the primitive actions. Empirically, we observe that on adding an optimal amount of controllability value in options, an agent optimizing for low variance in the TD error learns better than the one optimizing only for the cumulative reward. Intuitively, minimizing the variance in the TD error as a measure of safety helps the agents avoid states with high intrinsic variability in the reward. Depending on the nature of the game itself, we observe different degrees of response to different levels of controllability in Q*bert, Amidar, and MsPacman. Based on the amount of variability in the reward structure, each game benefits differently in the performance. Overall, a consistent boost in the performance is demonstrated with the proposed approach once the training is completed.

\begin{figure}[ht]
	\begin{center}
		\begin{subfigure}[b]{0.31\textwidth}
			\centering
			\captionsetup{justification=centering}
			\includegraphics[width=\textwidth]{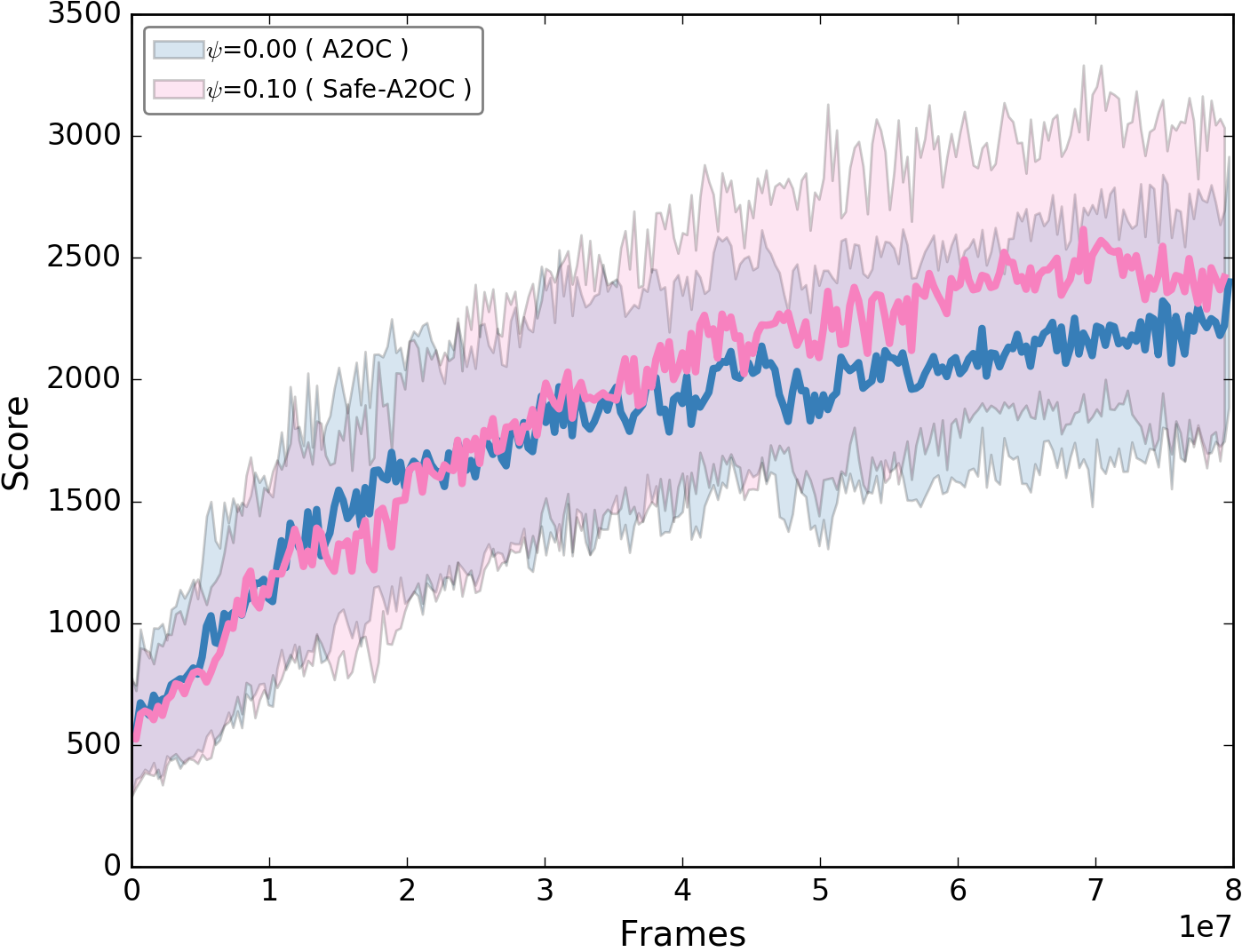}
			\caption[]%
			{{\small MsPacman}}    
			\label{fig:MsPacman4options}
		\end{subfigure}
		\begin{subfigure}[b]{0.31\textwidth}  
			\centering
			\captionsetup{justification=centering}
			\includegraphics[width=\textwidth]{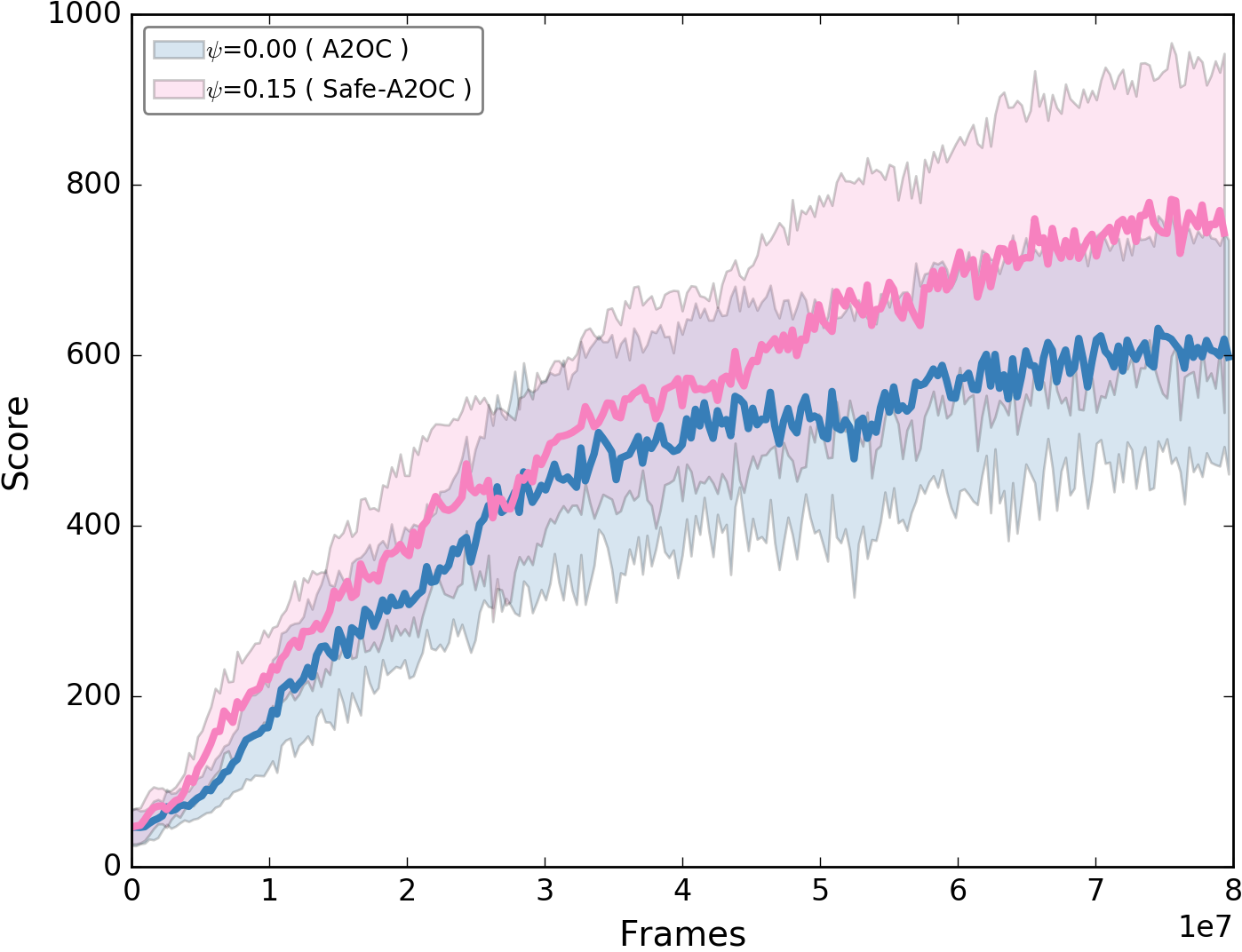}
			\caption[]%
			{{\small Amidar}}    
			\label{fig:Amidar4options}
		\end{subfigure}
		\begin{subfigure}[b]{0.31\textwidth}   
			\centering
			\captionsetup{justification=centering}
			\includegraphics[width=\textwidth]{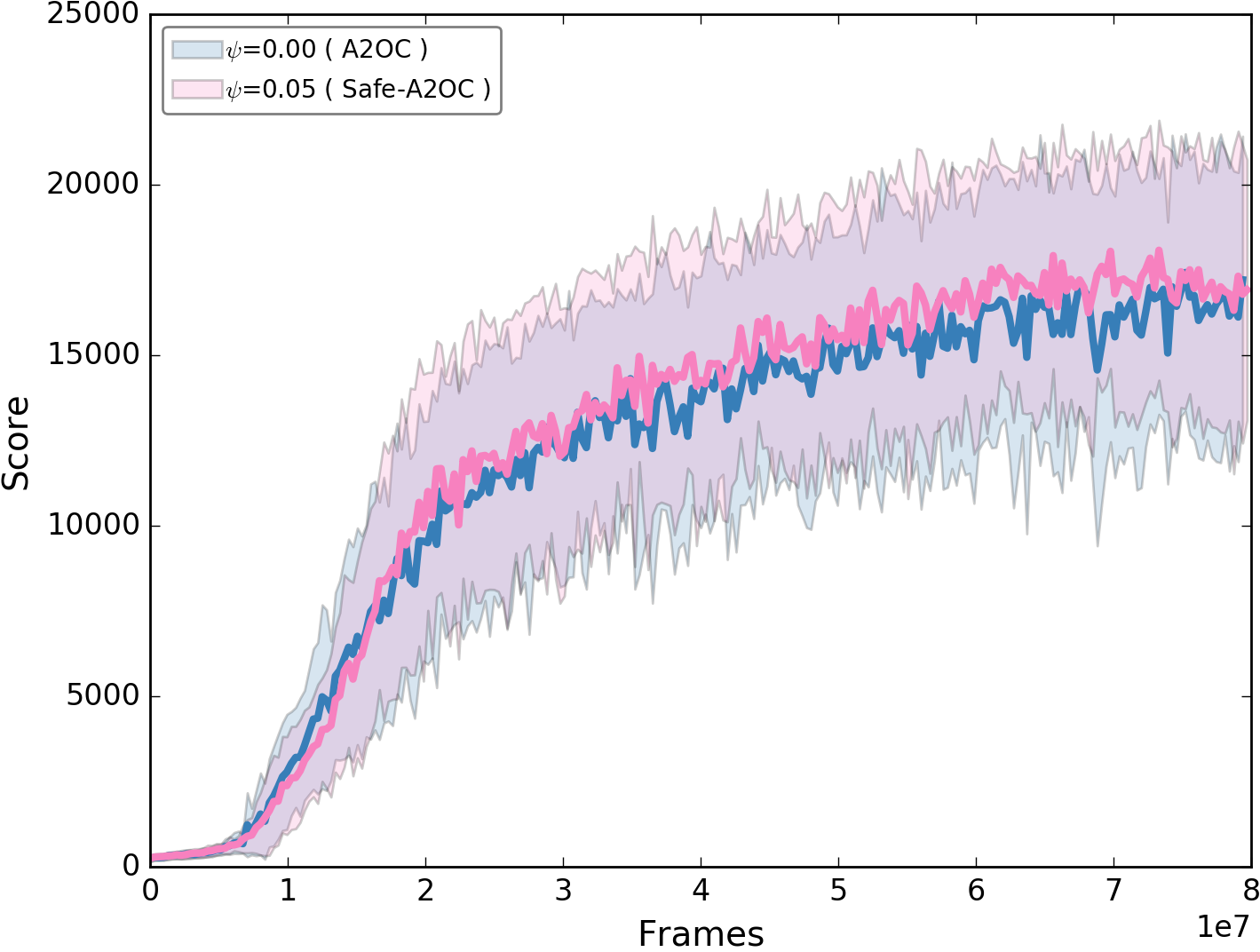}
			\caption[]%
			{{\small QBert}}    
			\label{fig:Qbert4options}
		\end{subfigure}
		\caption[]{\textbf{Learning curves in Atari games during training:} demonstrate that options with a controllability factor of $\psi=0.10$ learn better than the best performing A2OC baseline ($\psi=0$, $\epsilon=0.2$). Higher value of $\psi$ results in a poor performance. Performance is averaged across $5$ independent seeds for each game.} 
		\label{fig:ALELearning_Curves}
	\end{center}
\end{figure}

We also inspect the learning curves during the training phase for all three Atari games. Fig.~\ref{fig:ALELearning_Curves} depicts the learning curves over $80$M frames with varying controllability parameter. The results are averaged across $5$ runs per game. In Amidar we observe that with a specific degrees of controllability ($\psi = 0.15$), options learned with our notion of safety (Safe-A2OC) outperforms the baseline option-critic (A2OC). An improvement in the training performance also translates well to the testing phase as seen in the Table~\ref{tab:finalscores}. Interestingly, we observe that the performance of Safe-A2OC improves only marginally during the training phase in Ms-Pacman and is at-par with A2OC in Q*Bert. However, it is noteworthy that while we do not see much improvement in the training curves for both MsPacman and Q*Bert, during the testing phase, both these games show robust and improved performance. According to our interpretation, explicitly optimizing for the variance in TD error might not always result in a performance boost. Nevertheless, the proposed objective function makes the agent aware of intrinsic variability in the rewards. In the subsequent section, we will qualitatively analyze the behaviour of the trained agents.

\textbf{Qualitative Observations:} 
We are now interested in better understanding the behaviour of the trained agents. We observe that different values of $\psi$ control the degree to which an agent tends to be risk averse. A grid search over the different degrees of the controllability hyper parameter $\psi$ resulted in a narrow range of $0$ to $0.15$. For a very high value of $\psi>0.3$, we observe that the agents become extremely risk-averse resulting in a poor performance. An optimum value of $\psi$ for all three games is obtained around $0.05-0.15$. For a qualitative analysis, we also present the videos of the trained agents\footnote{\textbf{Videos} of trained agents in Atari games are available at \url{https://sites.google.com/view/safe-option-critic}.}.

\begin{figure}[t]
	\begin{center}
		\centerline{\includegraphics[width=\columnwidth]{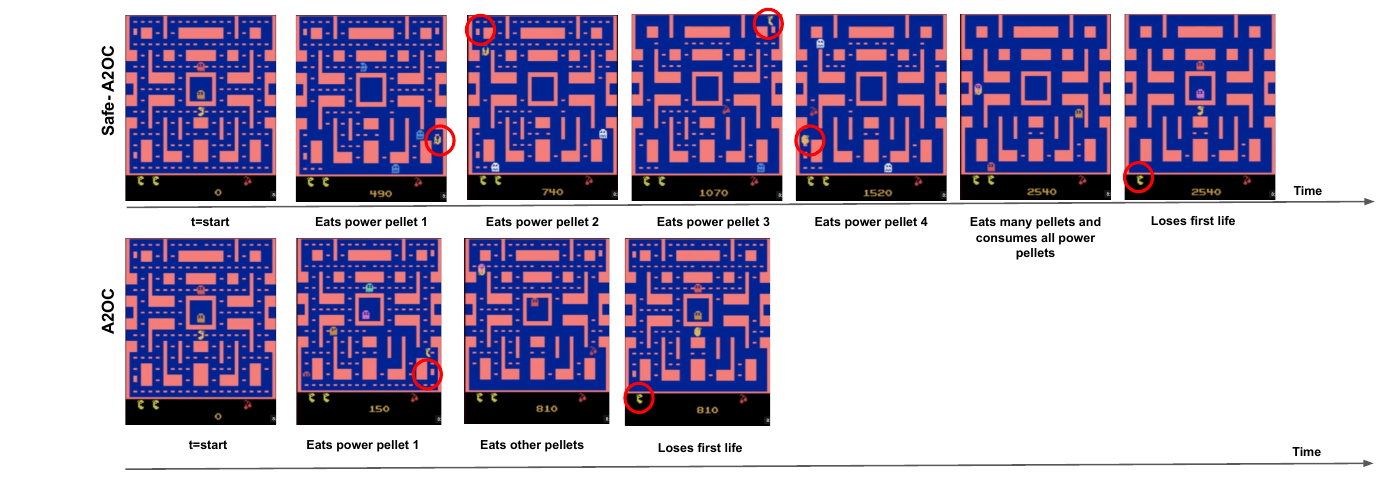}}
		\caption{\textbf{Qualitative analysis in MsPacman}: Our interpretation is that the Safe-A2OC agent understands the source of intrinsic variability in the reward due to explicitly optimizing for the variance in TD error. This results in Safe-A2OC agent to adopt the strategy of eating the large flashing \textit{power pellets}, causing the ghosts to flee. The Safe-A2OC agent is able to survive longer and ends up accumulating more cumulative reward as opposed to the baseline A2OC agent. Here, we depict a sampled trajectory of the Safe-A2OC agent in comparison with the baseline A2OC agent.}
		\label{fig:MsPacmananalysis}
	\end{center}
\end{figure}

Upon visual inspection, we observe improvements in game playing strategies for Safe-OC agent as compared to its counterpart OC. The Safe-A2OC agents demonstrate relatively better understanding of the source of intrinsic variability in the reward structure. For instance, consider the game of MsPacman. The agent is tasked with eating all of the pellets in an enclosed maze while avoiding the four colored ghosts. On consuming the power pellets in the environment, the ghosts turns blue and flee, resulting in bonus points. Frame-by-frame analysis shows that the Safe-A2OC agent is able to escape from the ghosts and stay alive longer when these ghosts are harmful in the context of a terminal state. Figure~\ref{fig:MsPacmananalysis} depicts this observation, where, the A2OC agent loses its first life relatively quickly as compared to the Safe-A2OC agent. These insights demonstrate that explicitly optimizing for the variance in TD error results in avoiding the visits to states with intrinsic variability in the rewards. Particularly, in MsPacman, the additional cost in the objective function helps the agent to understand the intrinsic variability in reward structure such as the behaviour observed upon the acquisition of the power pellets.

Furthermore, on close analysis of the behaviour of the Safe-OC agents, we observe agents with extremely high value of $\psi$ are relatively more risk-averse but less exploratory in nature, thereby resulting in a poor performance (see Appendix \ref{app:ale_exp}). This corroborates with the findings in the learning curves as shown in the  Fig.~\ref{fig:ALELearning_Curves}. For an optimal value of controllability regularizer ($\psi$), we observe a balanced amount of risk-aversion, and therefore, the agent adopts a behaviour which demonstrates a better understanding of intrinsic variability in the rewards structure. We perform a similar analysis for the game of Amidar and report similar insights in the Appendix~\ref{sec-amidarqualanalysis}.

\section{DISCUSSION}
\label{sec:conclusionandfuturework}
In this work, we introduced \textit{Safe Option-Critic}, a general framework that extends the idea of controllability from the primitive action space to the option-critic architecture. The key idea is to discourage the agent from visiting the state space with high uncertainty in their behavioural outcomes by constraining the variance in TD error. Recently, \cite{sherstan2018directly} proposed a direct method to estimate the variance of $\lambda$ return. The authors proposed a Bellman operator for the variance that uses the square of the TD error. This work further supports our approach to estimating the risk using the squared TD error.

Our experiments in the tabular case empirically demonstrate the reduced variance in the return. Moreover, we observe a boost in the overall performance for both the tabular and the linear function approximation. Experiments in the ALE domain demonstrate that the agent can learn about the intrinsic variability in a large and complicated state-space using non-linear function approximation. Qualitative insights from ALE also demonstrate that the options with a notion of safety are more cautious in their behaviour, and therefore, result in improved overall performance.

\textbf{Limitations and Future Work}: 
In this work, we limit the return calculation until an option terminates. Using the $n$-step returns at the SMDP level is of potential interest for future work. Additionally, in this work we assume that all the options are available in every state. In the context of safety, it would be of interest to understand the setting where options initiation sets are limited to a subset of the entire state space \citep{khetarpal2020options}. One could also add a scheduler on the controllability regularizer $\psi$ over time. For instance, $\psi$ could be initiated with a small value to encourage exploration in the initial stages of learning, and gradually the value of $\psi$ could be increased to limit the exploration in the unsafe states.

Besides, a more formal analysis of the agent behaviour in ALE can provide a rigorous understanding of the interplay between learning the dynamics and the controllability parameter. For instance, studying the training and policy traces by human subjects is scope for future work. 

In the proposed algorithm, we consider the notion of safety by accounting for the controllability of only the initial state-option pair. However, one can instead account for the variability due to the entire state-space by means of bootstrapping the target value function, aliasing due to the function approximation, and random resets to starting state. A potential direction for future work is to extend the controllability from the initial state-option pair to all the state-option pairs in the trajectory. This extension could potentially enhance the effects of the risk mitigation and speed up learning. The proposed notion of safety can also be extended to different levels of hierarchy, e.g. a mixture of options with varying degrees of controllability can be learned.

\section*{Acknowledgment}
The authors would like to thank Open Philanthropy for funding this work, Compute Canada for the computing resources, Herke van Hoof, Ayush Jain, Pierre-Luc Bacon, Gandharv Patil, Jean Harb, Martin Klissarov, Kushal Arora, for constructive discussions throughout the duration of this work, and the anonymous reviewers for the feedback on earlier drafts of this manuscript.

\newpage
\setcounter{section}{0}
\setcounter{figure}{0}
\section*{Appendix}

\subsection{\textbf{Experiments in the ALE Domain}}\label{app:ale_exp}
In this section, we show the results of training the agent in ALE games with multiple values of controllability parameter ($\psi$). The results are averaged across $5$ runs per game.

\begin{figure}[ht]
	\begin{center}
		\begin{subfigure}[b]{0.31\textwidth}
			\centering
			\captionsetup{justification=centering}
			\includegraphics[width=\textwidth]{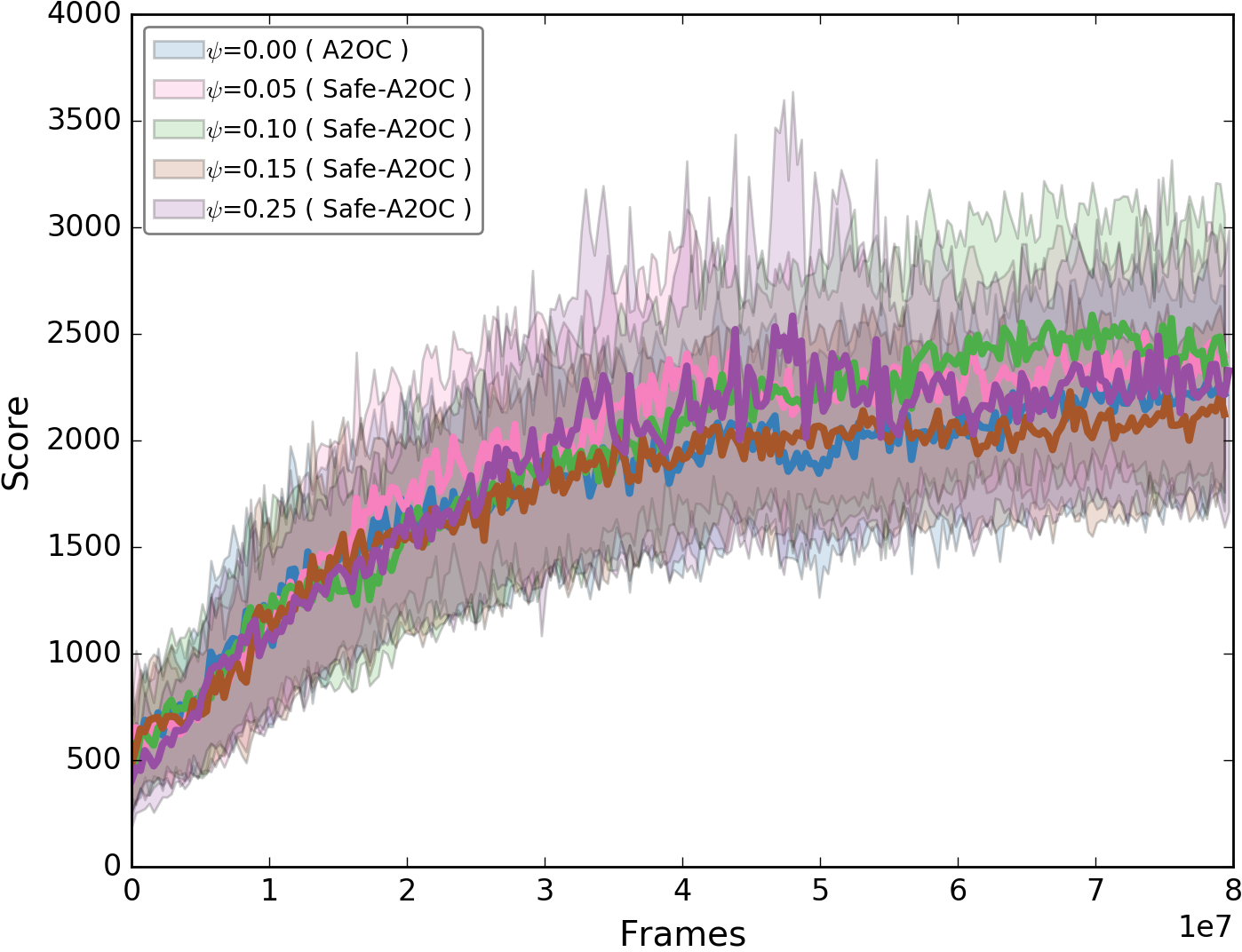}
			\caption[]%
			{{\small MsPacman}}    
			\label{fig:MsPacman4option}
		\end{subfigure}
		\begin{subfigure}[b]{0.31\textwidth}  
			\centering
			\captionsetup{justification=centering}
			\includegraphics[width=\textwidth]{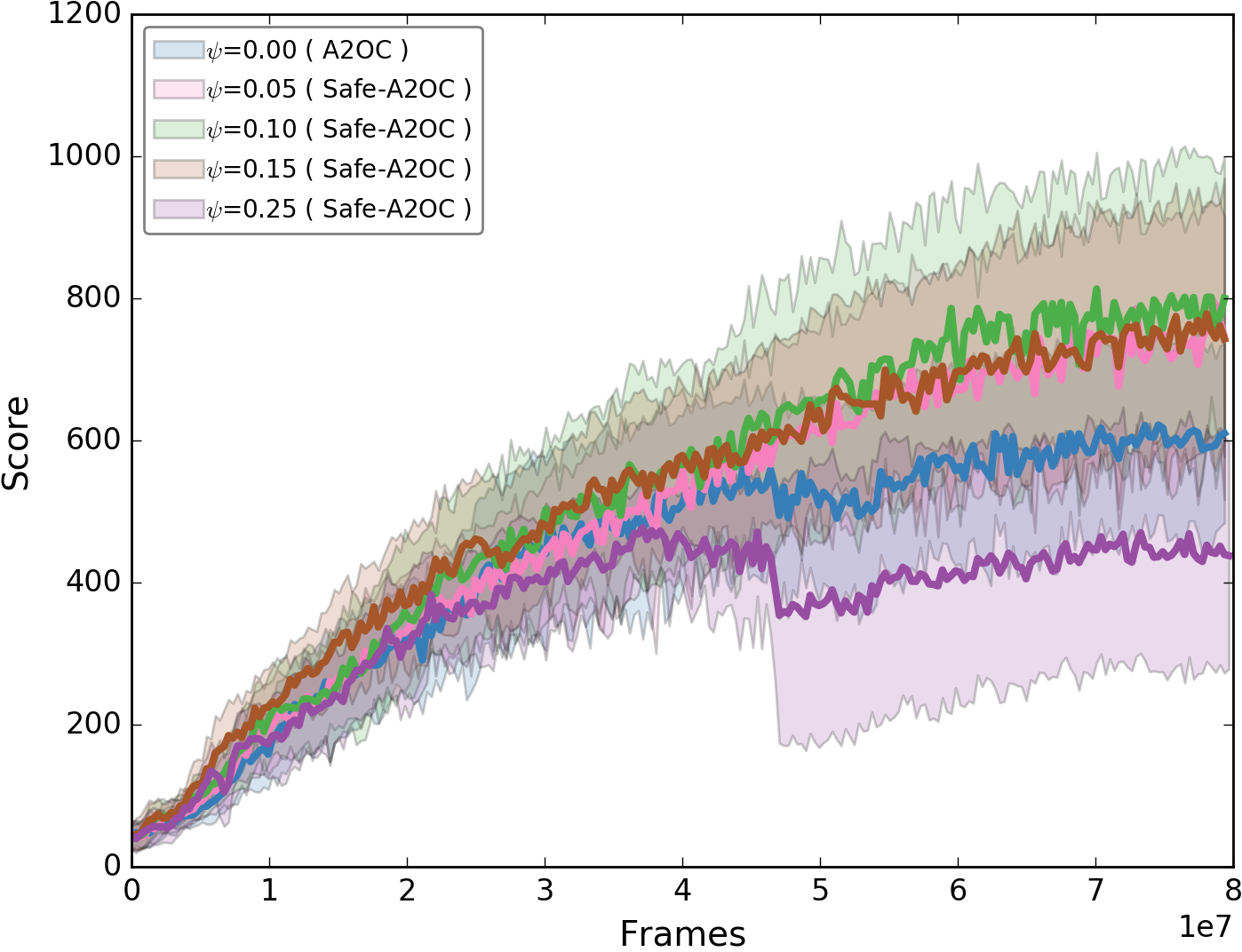}
			\caption[]%
			{{\small Amidar}}    
			\label{fig:Amidar4option}
		\end{subfigure}
		\begin{subfigure}[b]{0.31\textwidth}   
			\centering
			\captionsetup{justification=centering}
			\includegraphics[width=\textwidth]{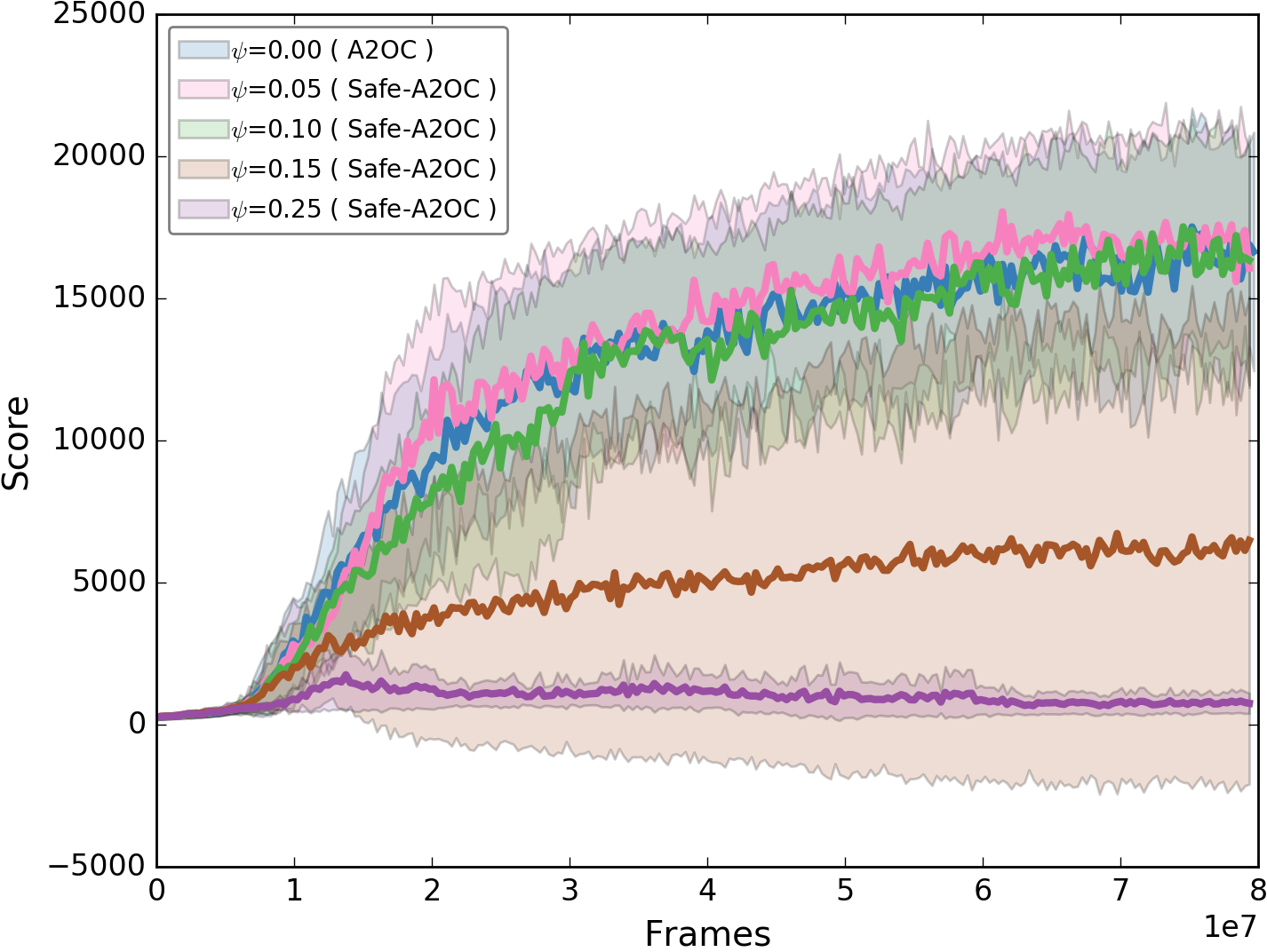}
			\caption[]%
			{{\small Q*Bert}}    
			\label{fig:Qbert4option}
		\end{subfigure}
		\caption[]{\textbf{Learning curves during training for ALE games:} Higher degrees of controllability ($\psi>0.15$) results in reduced exploration and adversely effects the performance. We observe that the agents trained with an extremely high value of $\psi$ are relatively more risk averse as compared to smaller values of controllability.} 
		\label{fig:ALELearningCurves}
	\end{center}
\end{figure}

\subsection{\textbf{Qualitative Observations in Amidar}}
\label{sec-amidarqualanalysis}
\begin{figure}[t]
	\begin{center}
		\centerline{\includegraphics[width=0.8\columnwidth]{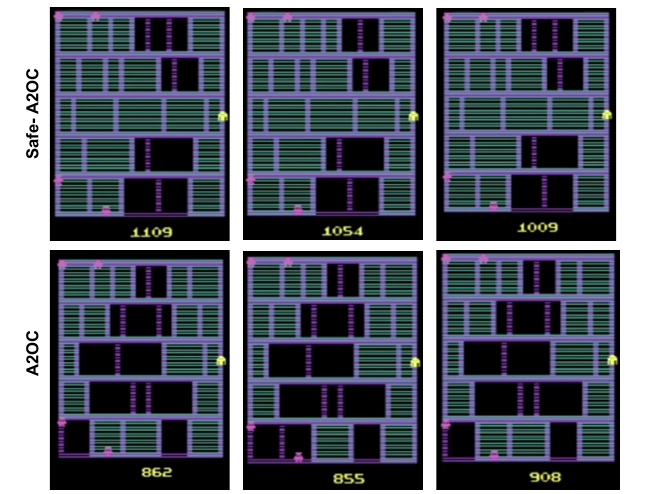}}
		\caption{\textbf{Qualitative analysis in Amidar}: We here demonstrate the final frames from randomly sampled 3 games played by the trained agents. Safe-A2OC agent learns to paint relatively more rectangles as compared to the A2OC agent as painting all sides of a rectangle results in making the ghosts ineffective, and thus increases the cumulative return. Amidar offers intrinsic variability in the reward structure and our approach empowers the agents to understand it better.}
		\label{fig:Amidaranalysis}
	\end{center}
\end{figure}

We also analyze the behaviour of the trained agents in the game of Amidar as shown in Figure~\ref{fig:Amidaranalysis}. Just as MsPacman, in Amidar, the agent is faced with enemies, who upon contact can kill the agent. The agent's goal is to paint the rectangular boards by traversing all the sides of the board. Upon painting all the four corners of a rectangle, the agent is briefly equipped with the ability to make the enemies ineffective for a short duration, analogous to consuming the power-pellets in MsPacman. 

Fig~\ref{fig:Amidaranalysis} depicts the final frame in 3 games played by the trained agents. Safe-A2OC agents end up painting more rectangles than the A2OC agent. There is inherent variability in the reward structure of this environment. Since painting all four sides of the rectangle results in the ability to make the enemies ineffective, Safe-A2OC agents are able to understand this structure relatively better by directly optimizing for the variance in TD error. Therefore, Safe-A2OC agents result in painting more rectangles in the board as depicted, leading to a higher cumulative reward than the baseline agent.


\begin{thebibliography}{}

\bibitem[Amodei et~al., 2016]{concreteproblems}
Amodei, D., Olah, C., Steinhardt, J., Christiano, P.~F., Schulman, J., and
  Man{\'{e}}, D. (2016).
\newblock Concrete problems in {AI} safety.
\newblock {\em CoRR}.

\bibitem[Bacon et~al., 2017]{bacon2017option}
Bacon, P.-L., Harb, J., and Precup, D. (2017).
\newblock The option-critic architecture.
\newblock In {\em AAAI}, pages 1726--1734.

\bibitem[Barreto et~al., 2019]{barreto2019option}
Barreto, A., Borsa, D., Hou, S., Comanici, G., Ayg{\"u}n, E., Hamel, P.,
  Toyama, D., Mourad, S., Silver, D., Precup, D., et~al. (2019).
\newblock The option keyboard: Combining skills in reinforcement learning.
\newblock In {\em Advances in Neural Information Processing Systems}, pages
  13052--13062.

\bibitem[Barto and Mahadevan, 2003]{barto2003recent}
Barto, A.~G. and Mahadevan, S. (2003).
\newblock Recent advances in hierarchical reinforcement learning.
\newblock {\em Discrete Event Dynamic Systems}, 13(4):341--379.

\bibitem[Bellemare et~al., 2013]{bellemare2013arcade}
Bellemare, M.~G., Naddaf, Y., Veness, J., and Bowling, M. (2013).
\newblock The arcade learning environment: An evaluation platform for general
  agents.
\newblock {\em Journal of Artificial Intelligence Research}, 47:253--279.

\bibitem[Borkar and Meyn, 2002]{borkar2002risk}
Borkar, V.~S. and Meyn, S.~P. (2002).
\newblock Risk-sensitive optimal control for {M}arkov decision processes with
  monotone cost.
\newblock {\em Mathematics of Operations Research}, 27(1):192--209.

\bibitem[Daniel et~al., 2016]{daniel2016probabilistic}
Daniel, C., Van~Hoof, H., Peters, J., and Neumann, G. (2016).
\newblock Probabilistic inference for determining options in reinforcement
  learning.
\newblock {\em Machine Learning}, 104(2-3):337--357.

\bibitem[Dietterich, 2000]{dietterich2000hierarchical}
Dietterich, T.~G. (2000).
\newblock Hierarchical reinforcement learning with the {MAXQ} value function
  decomposition.
\newblock {\em Journal of Artificial Intelligence Research}, 13:227--303.

\bibitem[Fikes et~al., 1972]{fikes1972learning}
Fikes, R.~E., Hart, P.~E., and Nilsson, N.~J. (1972).
\newblock Learning and executing generalized robot plans.
\newblock {\em Artificial Intelligence}, 3:251--288.

\bibitem[Fikes et~al., 1981]{fikes1981learning}
Fikes, R.~E., Hart, P.~E., and Nilsson, N.~J. (1981).
\newblock Learning and executing generalized robot plans.
\newblock In {\em Readings in Artificial Intelligence}, pages 231--249.
  Elsevier.

\bibitem[{Future of Life Institute}, 2017]{asilomar2017}
{Future of Life Institute} (2017).
\newblock {Asilomar AI Principles}.

\bibitem[Garc{\i}a and Fern{\'a}ndez, 2015]{garcia2015comprehensive}
Garc{\i}a, J. and Fern{\'a}ndez, F. (2015).
\newblock A comprehensive survey on safe reinforcement learning.
\newblock {\em Journal of Machine Learning Research}, 16(1):1437--1480.

\bibitem[Gehring and Precup, 2013]{Gehring:2013}
Gehring, C. and Precup, D. (2013).
\newblock Smart exploration in reinforcement learning using absolute temporal
  difference errors.
\newblock In {\em Proceedings of the 2013 International Conference on
  Autonomous Agents and Multi-agent Systems}, AAMAS '13, pages 1037--1044.

\bibitem[Geibel and Wysotzki, 2005]{geibel2005risk}
Geibel, P. and Wysotzki, F. (2005).
\newblock Risk-sensitive reinforcement learning applied to control under
  constraints.
\newblock {\em J. Artif. Intell. Res.(JAIR)}, 24:81--108.

\bibitem[Harb et~al., 2018]{harb2017waiting}
Harb, J., Bacon, P.-L., Klissarov, M., and Precup, D. (2018).
\newblock When waiting is not an option: Learning options with a deliberation
  cost.
\newblock In {\em AAAI}.

\bibitem[Heger, 1994]{heger1994consideration}
Heger, M. (1994).
\newblock Consideration of risk in reinforcement learning.
\newblock In {\em Machine Learning Proceedings 1994}, pages 105--111. Elsevier.

\bibitem[Howard and Matheson, 1972]{howard1972risk}
Howard, R.~A. and Matheson, J.~E. (1972).
\newblock Risk-sensitive {M}arkov decision processes.
\newblock {\em Management Science}, 18(7):356--369.

\bibitem[Iba, 1989]{iba1989heuristic}
Iba, G.~A. (1989).
\newblock A heuristic approach to the discovery of macro-operators.
\newblock {\em Machine Learning}, 3(4):285--317.

\bibitem[Iyengar, 2005]{iyengar2005robust}
Iyengar, G.~N. (2005).
\newblock Robust dynamic programming.
\newblock {\em Mathematics of Operations Research}, 30(2):257--280.

\bibitem[Jain et~al., 2021]{jain2021variance}
Jain, A., Patil, G., Jain, A., Khetarpal, K., and Precup, D. (2021).
\newblock Variance penalized on-policy and off-policy actor-critic.
\newblock {\em arXiv preprint arXiv:2102.01985}.

\bibitem[Jain and Precup, 2018]{jain2018eligibility}
Jain, A. and Precup, D. (2018).
\newblock Eligibility traces for options.
\newblock In {\em Proceedings of the 17th International Conference on
  Autonomous Agents and MultiAgent Systems}, pages 1008--1016.

\bibitem[Khetarpal et~al., 2020]{khetarpal2020options}
Khetarpal, K., Klissarov, M., Chevalier-Boisvert, M., Bacon, P.-L., and Precup,
  D. (2020).
\newblock Options of interest: Temporal abstraction with interest functions.
\newblock In {\em Proceedings of the AAAI Conference on Artificial
  Intelligence}, volume~34, pages 4444--4451.

\bibitem[Konidaris and Barto, 2007]{konidaris2007building}
Konidaris, G. and Barto, A.~G. (2007).
\newblock Building portable options: Skill transfer in reinforcement learning.
\newblock In {\em IJCAI}, volume~7, pages 895--900.

\bibitem[Konidaris et~al., 2011]{konidaris2011autonomous}
Konidaris, G., Kuindersma, S., Grupen, R.~A., and Barto, A.~G. (2011).
\newblock Autonomous skill acquisition on a mobile manipulator.
\newblock In {\em AAAI}.

\bibitem[Korf, 1983]{korf1983learning}
Korf, R.~E. (1983).
\newblock {\em Learning to Solve Problems by Searching for Macro-operators}.
\newblock PhD thesis, Pittsburgh, PA, USA.
\newblock AAI8425820.

\bibitem[Kulkarni et~al., 2016]{kulkarni2016hierarchical}
Kulkarni, T.~D., Narasimhan, K., Saeedi, A., and Tenenbaum, J. (2016).
\newblock Hierarchical deep reinforcement learning: Integrating temporal
  abstraction and intrinsic motivation.
\newblock In {\em Advances in Neural Information Processing Systems}, pages
  3675--3683.

\bibitem[Law et~al., 2005]{law2005risk}
Law, E.~L., Coggan, M., Precup, D., and Ratitch, B. (2005).
\newblock Risk-directed exploration in reinforcement learning.
\newblock {\em Planning and Learning in A Priori Unknown or Dynamic Domains},
  page~97.

\bibitem[Lim et~al., 2013]{lim2013reinforcement}
Lim, S.~H., Xu, H., and Mannor, S. (2013).
\newblock Reinforcement learning in robust {M}arkov decision processes.
\newblock {\em Advances in Neural Information Processing Systems}, 26:701--709.

\bibitem[{Machado} et~al., 2017]{MarlosEvaluationProtocol}
{Machado}, M.~C., {Bellemare}, M.~G., {Talvitie}, E., {Veness}, J.,
  {Hausknecht}, M., and {Bowling}, M. (2017).
\newblock {Revisiting the Arcade Learning Environment: Evaluation Protocols and
  Open Problems for General Agents}.
\newblock {\em ArXiv e-prints}.

\bibitem[Mankowitz et~al., 2016]{mankowitz2016adaptive}
Mankowitz, D.~J., Mann, T.~A., and Mannor, S. (2016).
\newblock Adaptive skills adaptive partitions (asap).
\newblock In {\em Advances in Neural Information Processing Systems}, pages
  1588--1596.

\bibitem[McGovern and Barto, 2001]{mcgovern2001automatic}
McGovern, A. and Barto, A.~G. (2001).
\newblock Automatic discovery of subgoals in reinforcement learning using
  diverse density.
\newblock In {\em ICML}, volume~1, pages 361--368.

\bibitem[Menache et~al., 2002]{menache2002q}
Menache, I., Mannor, S., and Shimkin, N. (2002).
\newblock {Q}-cut - dynamic discovery of sub-goals in reinforcement learning.
\newblock In {\em European Conference on Machine Learning}, pages 295--306.
  Springer.

\bibitem[Mnih et~al., 2016]{mnih2016asynchronous}
Mnih, V., Badia, A.~P., Mirza, M., Graves, A., Lillicrap, T., Harley, T.,
  Silver, D., and Kavukcuoglu, K. (2016).
\newblock Asynchronous methods for deep reinforcement learning.
\newblock In {\em International Conference on Machine Learning}, pages
  1928--1937.

\bibitem[Nair et~al., 2015]{Nair}
Nair, A., Srinivasan, P., Blackwell, S., Alcicek, C., Fearon, R., Maria, A.~D.,
  Panneershelvam, V., Suleyman, M., Beattie, C., Petersen, S., Legg, S., Mnih,
  V., Kavukcuoglu, K., and Silver, D. (2015).
\newblock Massively parallel methods for deep reinforcement learning.
\newblock {\em CoRR}.

\bibitem[Nilim and El~Ghaoui, 2005]{nilim2005robust}
Nilim, A. and El~Ghaoui, L. (2005).
\newblock Robust control of {M}arkov decision processes with uncertain
  transition matrices.
\newblock {\em Operations Research}, 53(5):780--798.

\bibitem[Parr and Russell, 1998]{parr1998reinforcement}
Parr, R. and Russell, S.~J. (1998).
\newblock Reinforcement learning with hierarchies of machines.
\newblock In {\em Advances in neural information processing systems}, pages
  1043--1049.

\bibitem[Precup, 2000]{precup2000temporal}
Precup, D. (2000).
\newblock {\em Temporal abstraction in reinforcement learning}.
\newblock University of Massachusetts Amherst.

\bibitem[Riemer et~al., 2018]{riemer2018learning}
Riemer, M., Liu, M., and Tesauro, G. (2018).
\newblock Learning abstract options.
\newblock In {\em Advances in Neural Information Processing Systems}, pages
  10424--10434.

\bibitem[Sherstan et~al., 2018]{sherstan2018directly}
Sherstan, C., Ashley, D.~R., Bennett, B., Young, K., White, A., White, M., and
  Sutton, R.~S. (2018).
\newblock Comparing direct and indirect temporal-difference methods for
  estimating the variance of the return.
\newblock In {\em Proceedings of Uncertainty in Artificial Intelligence}, pages
  63--72.

\bibitem[Stolle and Precup, 2002]{stolle2002learning}
Stolle, M. and Precup, D. (2002).
\newblock Learning options in reinforcement learning.
\newblock In {\em International Symposium on abstraction, reformulation, and
  approximation}, pages 212--223. Springer.

\bibitem[Sutton, 1988]{sutton1988learning}
Sutton, R.~S. (1988).
\newblock Learning to predict by the methods of temporal differences.
\newblock {\em Machine Learning}, 3(1):9--44.

\bibitem[Sutton and Barto, 1998]{Sutton:1998:IRL:551283}
Sutton, R.~S. and Barto, A.~G. (1998).
\newblock {\em Introduction to Reinforcement Learning}.
\newblock MIT Press, Cambridge, MA, USA, 1st edition.

\bibitem[Sutton et~al., 2000]{sutton2000policy}
Sutton, R.~S., McAllester, D.~A., Singh, S.~P., and Mansour, Y. (2000).
\newblock Policy gradient methods for reinforcement learning with function
  approximation.
\newblock In {\em Advances in Neural Information Processing Systems}, pages
  1057--1063.

\bibitem[Sutton et~al., 1999]{SUTTON1999181}
Sutton, R.~S., Precup, D., and Singh, S. (1999).
\newblock Between {MDP}s and semi-{MDP}s: A framework for temporal abstraction
  in reinforcement learning.
\newblock {\em Artificial Intelligence}, 112(1-2):181--211.

\bibitem[Tamar et~al., 2012]{tamar2012policy}
Tamar, A., Di~Castro, D., and Mannor, S. (2012).
\newblock Policy gradients with variance related risk criteria.
\newblock In {\em Proceedings of the twenty-ninth International Conference on
  Machine Learning}, pages 387--396.

\bibitem[Tamar et~al., 2016]{tamar2016learning}
Tamar, A., Di~Castro, D., and Mannor, S. (2016).
\newblock Learning the variance of the reward-to-go.
\newblock {\em Journal of Machine Learning Research}, 17(13):1--36.

\bibitem[Tamar et~al., 2013]{tamar2013scaling}
Tamar, A., Xu, H., and Mannor, S. (2013).
\newblock Scaling up robust {MDP}s by reinforcement learning.
\newblock {\em arXiv preprint arXiv:1306.6189}.

\bibitem[Van~Hasselt et~al., 2016]{van2016deep}
Van~Hasselt, H., Guez, A., and Silver, D. (2016).
\newblock Deep reinforcement learning with double {Q}-learning.
\newblock In {\em AAAI}, volume~16, pages 2094--2100.

\bibitem[Vezhnevets et~al., 2016]{vezhnevets2016strategic}
Vezhnevets, A., Mnih, V., Osindero, S., Graves, A., Vinyals, O., Agapiou, J.,
  et~al. (2016).
\newblock Strategic attentive writer for learning macro-actions.
\newblock In {\em Advances in Neural Information Processing Systems}, pages
  3486--3494.

\bibitem[Wang et~al., 2015]{WangFL15}
Wang, Z., de~Freitas, N., and Lanctot, M. (2015).
\newblock Dueling network architectures for deep reinforcement learning.
\newblock {\em CoRR}.

\bibitem[White, 1994]{white1994mathematical}
White, D. (1994).
\newblock A mathematical programming approach to a problem in variance
  penalised {M}arkov decision processes.
\newblock {\em Operations-Research-Spektrum}, 15(4):225--230.

\end{thebibliography}
\end{document}